\documentclass{article}
\usepackage{arxiv}
\usepackage{natbib}

\usepackage{hyperref}
\usepackage{longtable} 
\usepackage{booktabs}
\usepackage{graphicx}
\usepackage{amsmath}
\usepackage{amssymb}
\usepackage{amsthm}
\theoremstyle{plain}
\newtheorem{theorem}{Theorem}[section]

\theoremstyle{definition}

\theoremstyle{remark}

\title{Adapting AI Weather Foundation Models to a Changing Climate}

\author{%
  \normalfont\large
  Xinlei Xiong\textsuperscript{1} \quad
  Wenbo Hu\textsuperscript{1} \quad
  Shuxun Zhou\textsuperscript{1}\\
  \normalfont\large
  Lingxi Xie\textsuperscript{2} \quad
  Richang Hong\textsuperscript{1}\quad Qi Tian\textsuperscript{2}  \\[0.7em]
  \normalfont\small
  \textsuperscript{1}Hefei University of Technology\\
  \normalfont\small
  \textsuperscript{2}Huawei Inc.
}
\date{}

\begin{document}

\maketitle

\begin{abstract}
AI weather foundation models learn from historical archives, while the climate distribution encountered during deployment continues to evolve. This growing model--climate mismatch motivates continual adaptation without repeated full-model retraining, but the structure of weather forecast errors determines how such adaptation should be performed. We develop an online adaptation framework that exploits two complementary error structures: geographically persistent biases and flow-dependent errors that vary with the atmospheric state. Analysis of Aurora-1.3B reveals strongly concentrated forecast-loss sensitivity in its decoder, motivating flow-dependent corrections within fixed low-dimensional sensitivity subspaces. A residual memory tracks persistent errors remaining after this correction, while a variable-wise online gate controls the contribution of the persistent component. We evaluate Aurora-1.3B using ERA5 from 1959--1968 for offline pretraining and a strictly chronological 1969--2022 stream for online evaluation, with updates restricted to previously available verification targets. For 2-metre temperature, distributional departure from the historical reference, measured directly from ERA5, is associated with larger frozen-model errors and greater adaptation gains, while the adapted model maintains more stable skill across evaluation periods. Across 69 output fields and five lead times from 6 to 72 hours, our method reduces MSE and MAE by 4.17\% and 2.76\%, respectively, relative to $\delta$-Adapter, the strongest of six online adaptation baselines. 
\end{abstract}

\section{Introduction}
AI-based weather forecasting has advanced rapidly, with models such as Pangu-Weather, GraphCast, and Aurora achieving competitive global medium-range forecasting skill \citep{bi2023accurate,lam2023learning,bodnar2025foundation}. Long-term deployment, however, poses a different challenge: these models learn from finite historical archives, whereas the distribution of weather continues to evolve as the climate changes \citep{sippel2020climate}. As deployment conditions depart from those represented during training, a frozen model can face increasing model--climate mismatch, while repeatedly retraining billion-parameter weather foundation models is computationally expensive. Related work on climate emulation has similarly highlighted the difficulty of generalizing across changing climate distributions \citep{stanley2026no}. This motivates continual adaptation from newly verified forecasts while restricting the scope and cost of each update.

Evolving climate distributions create the need for continual adaptation, while weather-specific forecast-error structure guides how that adaptation should be performed. Existing online time-series methods track changing or recurring patterns under sequentially revealed supervision \citep{gama2014survey,pham2022learning,zhao2025proactive}, but weather forecasting involves high-dimensional, spatially coupled fields with characteristic error structures. Persistent regional biases, particularly in near-surface variables such as 2-metre temperature (T2M), can remain coherent across successive forecast cycles \citep{libonati2008correction,haiden2018addressing}, whereas other errors vary strongly with the instantaneous atmospheric circulation \citep{gelaro1998sensitivity,leutbecher2008ensemble,ferranti2015flow}. These complementary structures suggest retaining verified residual information where errors persist while learning state-dependent corrections where errors vary with the flow. Because their relative importance differs across atmospheric variables and forecast conditions, an effective adaptation mechanism should also adjust their contribution online.

The main difficulty is making flow-dependent adaptation efficient for a billion-parameter model operating on high-dimensional atmospheric states. Classical predictability studies show that forecast-error growth can concentrate within relatively low-dimensional dynamically sensitive directions \citep{gelaro1998sensitivity,patil2001local,trevisan2004assimilation}. We observe an analogous structure in Aurora: historical forecast-loss gradients exhibit strongly concentrated and asymmetric decoder sensitivity, with rank-32 input-side subspaces in the later decoder stage capturing more than 90\% of the gradient trace at the median across layers (Figure~\ref{fig:svd}). Moreover, although these sensitivity bases are estimated offline and kept fixed during deployment, their activation changes with the evolving atmospheric circulation (Figure~\ref{fig:flow_activation}). This suggests that online adaptation can be confined to historically error-sensitive directions without losing responsiveness to the current atmospheric state.

Guided by these observations, we develop a structure-aware online adaptation framework for weather foundation models. Flow-dependent corrections are learned by updating coefficients within fixed low-dimensional forecast-sensitivity subspaces while the pretrained backbone remains frozen. An exponentially weighted spatial residual memory tracks persistent errors remaining after the flow-dependent correction, and a variable-wise online gate controls how much of this persistent correction is added to the forecast. The gate is updated from sequentially revealed verification losses, and we establish a regret guarantee against the better fixed correction choice in hindsight. All updates use only information available at each point in the chronological stream, preserving a strictly causal online protocol.

We evaluate Aurora-1.3B using ERA5 reanalysis \citep{hersbach2020era5}, with 1959--1968 used for offline pretraining and sensitivity estimation and 1969--2022 forming a strictly chronological 54-year online evaluation stream. To characterize changing deployment conditions independently of forecast performance, we measure the departure of the T2M distribution from the historical reference directly from ERA5 using Jensen--Shannon distance. Larger departures are associated with higher frozen-model errors and greater adaptation gains, while the adapted model maintains more stable T2M skill across evaluation periods (Figure~\ref{fig:t2m_changing_climate}). Across 69 output fields and lead times of 6, 12, 24, 48, and 72 hours, our method reduces MSE and MAE by 4.17\% and 2.76\%, respectively, relative to $\delta$-Adapter~\citep{liang2026deltaadapter}, the strongest of six online adaptation baselines. Together, these results support continual adaptation as deployment climates evolve and show that forecast-error structure provides a principled basis for how that adaptation can be carried out efficiently.
\section{Related Works}

\textbf{AI weather forecasting models.}
Data-driven global weather forecasting has advanced rapidly through models such as FourCastNet~\citep{kurth2023fourcastnet}, Pangu-Weather~\citep{bi2023accurate}, GraphCast~\citep{lam2023learning}, FuXi~\citep{chen2023fuxi}, Aurora~\citep{bodnar2025foundation}, GenCast~\citep{price2025probabilistic}, and AIFS~\citep{lang2024aifs}. These systems achieve competitive deterministic or probabilistic medium-range forecasting, often at substantially lower inference cost than conventional numerical weather prediction. Probabilistic weather forecasting has also begun to distinguish uncertainty associated with atmospheric-state perturbations from uncertainty in learned model parameters, as explored in Pangu-Bayes~\citep{hu2025resolving}. Most existing forecasting systems, however, are trained offline and deployed with fixed parameters. Continual adaptation from sequentially verified forecasts as deployment distributions evolve remains comparatively underexplored.

\textbf{Online forecasting under distribution shift.}
General time-series forecasting has also explored informative temporal scales~\citep{liu2026ssformer} and cross-variable dependencies~\citep{gao2023client}, primarily through forecasting architectures rather than sequential verification-driven updates. Online time-series methods adapt predictors as new targets become available under non-stationary dynamics. Representative approaches include FSNet~\citep{pham2022learning}, OneNet~\citep{zhang2023onenet}, Proceed~\citep{zhao2025proactive}, $\delta$-Adapter~\citep{liang2026deltaadapter}, ADAPT-Z~\citep{huang2026adaptz}, and ELF~\citep{elf2025}. These methods address distribution shift through parameter updates, lightweight corrections, feature adaptation, memory, or online ensembling. Weather forecasting, however, exhibits persistent spatial biases and rapidly varying flow-dependent errors that call for different forms of correction. Our approach explicitly models these complementary structures and adapts their relative contribution online.

\textbf{Weather forecast-error structure.}
Persistent and flow-dependent forecast errors have long been studied in numerical weather prediction. Sequential bias-correction methods use previously verified residuals to reduce systematic regional errors~\citep{libonati2008correction,haiden2018addressing}, while predictability studies show that forecast-error growth depends on the instantaneous circulation and can concentrate along dynamically sensitive directions~\citep{gelaro1998sensitivity,patil2001local,leutbecher2008ensemble,ferranti2015flow,trevisan2004assimilation}. Neural weather forecasting has also begun to explore perturbation generation and model sensitivity~\citep{scher2021ensemble,tian2026evaluating}. Related reliability research has addressed calibrated predictive intervals through maximum mean discrepancy~\citep{cui2020calibrated} and uncertainty-aware time-series imputation through quantile sub-ensembles~\citep{liu2023subensembles}. These approaches target predictive uncertainty or missing-data recovery, whereas our focus is the online correction of structured deterministic forecast errors. We build on the meteorological insights above by using forecast-error structure not merely for diagnosis or ensemble generation, but to determine how a weather foundation model should adapt online.

\begin{figure}[t]
    \centering
    \includegraphics[width=0.80\linewidth]{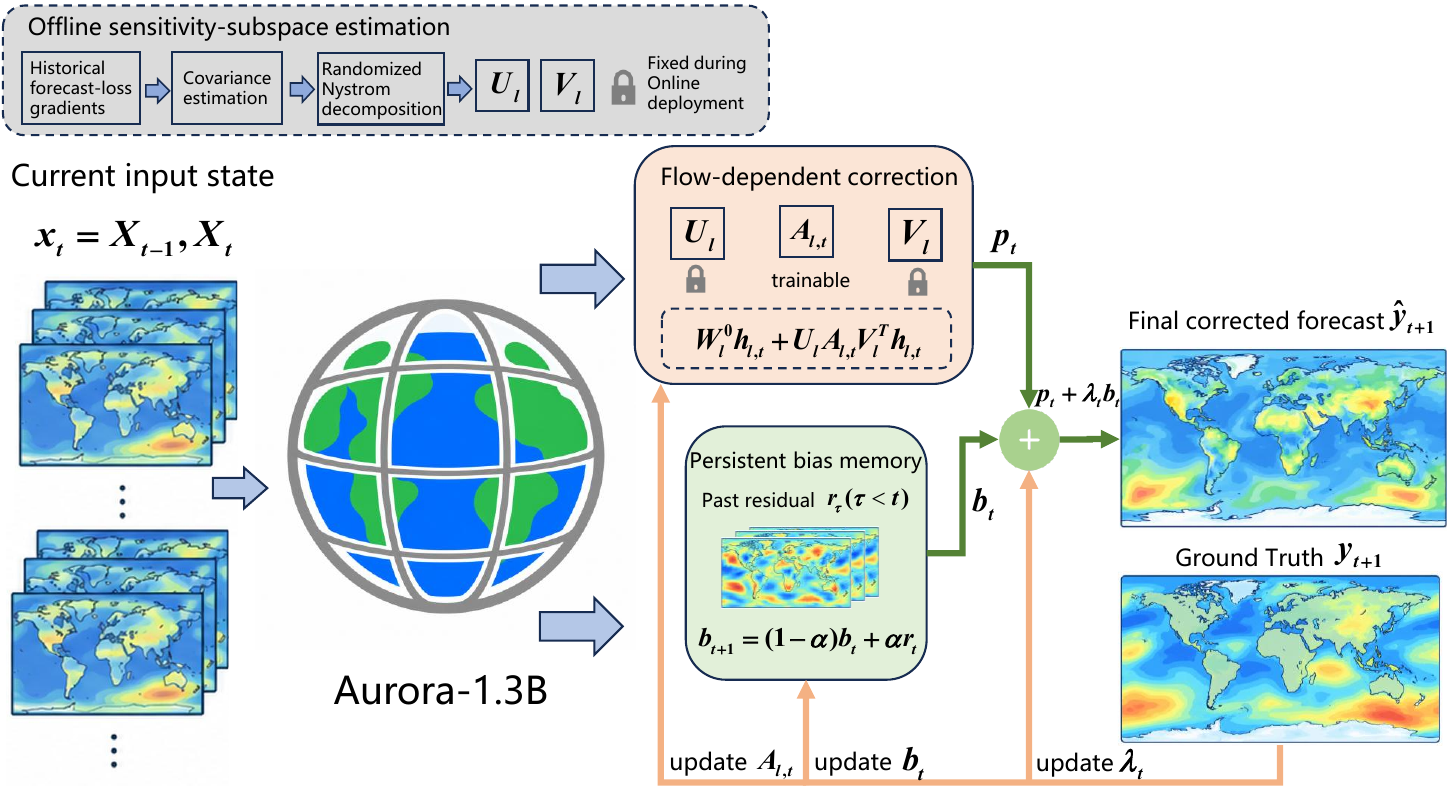}
    \caption{\textbf{Overview of the proposed online adaptation framework.}
    The Aurora-1.3B backbone remains frozen, while flow-dependent corrections are learned within fixed forecast-sensitivity subspaces and persistent residual errors are tracked by an online residual memory. A variable-wise gate adaptively controls the contribution of the persistent residual correction, and all updates use only previously revealed verification.}
    \label{fig:framework}
\end{figure}
\section{Methods}
We consider supervised online adaptation of an AI weather foundation model (AI-WFM) under a strictly chronological data stream. We adopt Aurora~\citep{bodnar2025foundation}, an Earth-system forecasting model with approximately 1.3 billion parameters, as the backbone. Our framework is guided by the structure of forecast errors encountered during deployment. We first characterize persistent and flow-dependent error structures that motivate complementary forms of correction. We then estimate low-dimensional forecast-sensitivity subspaces from historical data to enable efficient flow-dependent adaptation. Finally, a spatial residual memory tracks persistent errors remaining after the flow-dependent correction, while a variable-wise online gate controls the contribution of this persistent correction. Figure~\ref{fig:framework} provides an overview of the framework.

\subsection{Structured Forecast Errors in Long-Term Deployment}
\label{sec:method_overview}

Forecast errors encountered during long-term deployment exhibit complementary temporal and spatial structures. Some errors are geographically coherent and persist across successive forecast cycles, particularly for near-surface variables~\citep{libonati2008correction}. Other errors depend strongly on the instantaneous atmospheric circulation, with their magnitude and spatial structure varying with the evolving flow~\citep{leutbecher2008ensemble,ferranti2015flow}. These structures motivate different forms of online correction: persistent errors can be tracked using previously verified residuals, whereas flow-dependent errors require corrections that respond to the current atmospheric state.

Consider online forecasting round $t$. Let $x_t=(X_{t-1},X_t)$ denote the multivariate atmospheric input state, $y_{t+1,c}(s)$ the verifying target, and $f_{\theta_0,c}(x_t;s)$ the corresponding prediction of the frozen foundation model. Here, $\theta_0$ denotes the fixed pretrained parameters, $c$ indexes an atmospheric variable, and $s$ denotes its spatial coordinate. We operationally represent the forecast error as
\begin{equation}
e_{t,c}(s)
=
y_{t+1,c}(s)-f_{\theta_0,c}(x_t;s)
=
b_{t,c}(s)
+d_{t,c}(x_t;s)
+\varepsilon_{t,c}(s).
\label{eq:structured_error_decomposition}
\end{equation}
The term $b_{t,c}(s)$ represents a temporally persistent and spatially coherent component, while $d_{t,c}(x_t;s)$ denotes a state-dependent, flow-responsive component. The remaining term $\varepsilon_{t,c}(s)$ collects forecast errors that cannot be reliably captured from the available online information. Equation~\ref{eq:structured_error_decomposition} is an operational decomposition for adaptation rather than a causal decomposition of climate-change effects, and the components need not be strictly orthogonal. We next construct a compact model-space representation for the flow-dependent component.

\subsection{Forecast-Sensitivity Subspaces for Flow-Dependent Adaptation}
\label{sec:gradient_subspace}

\begin{figure}
    \centering
    \includegraphics[width=0.9\linewidth]{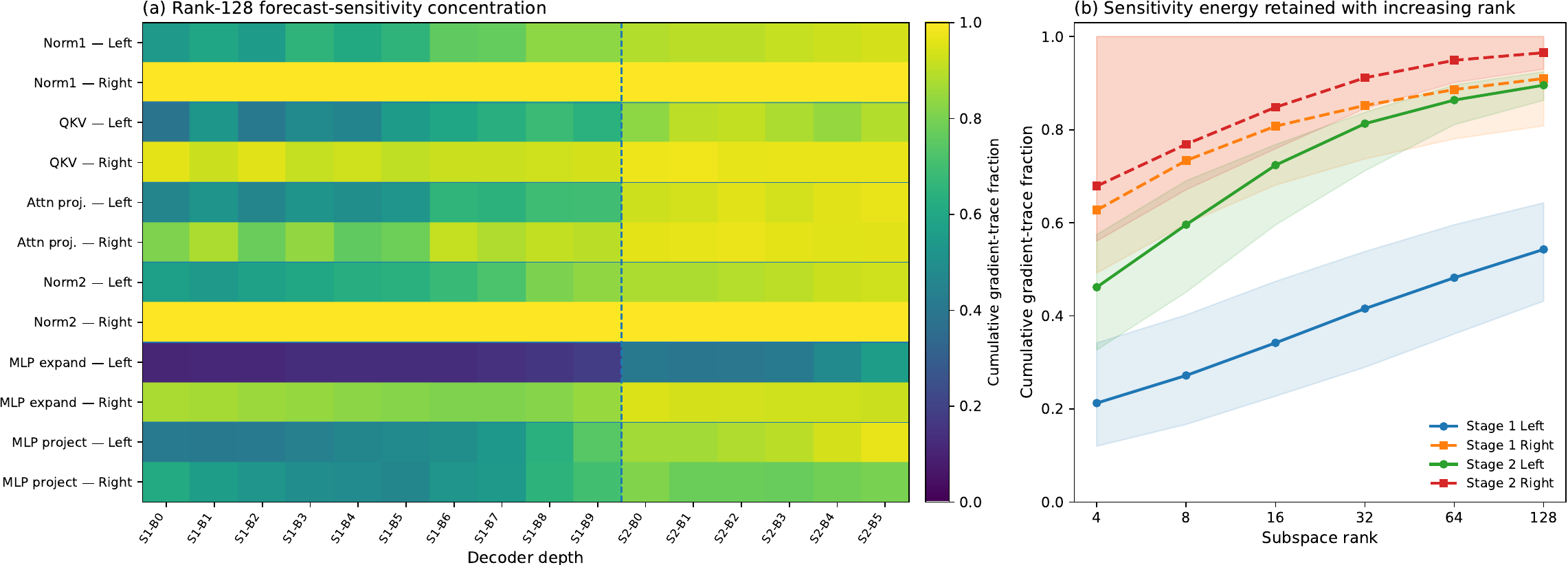}
    \caption{\textbf{Low-dimensional structure of forecast-loss sensitivity in the Aurora decoder.} \textbf{(a)} Cumulative gradient-trace fraction captured by the leading rank-$128$ left and right sensitivity subspaces across decoder layers. \textbf{(b)} Cumulative gradient-trace fraction versus subspace rank $r$; lines show the median across layers and shaded regions denote the interquartile range.}
    \label{fig:svd}
\end{figure}

NWP predictability studies have shown that, despite the high dimensionality of atmospheric state space, dominant finite-time forecast-error growth can concentrate within a small number of dynamically sensitive directions~\citep{gelaro1998sensitivity,patil2001local,annan2004orthogonality}. Motivated by this structure, we examine whether forecast-loss sensitivity within a weather foundation model is similarly concentrated in a low-dimensional subspace.

For each selected linear layer $\ell$, we characterize historical forecast-loss sensitivity using per-sample weight gradients
\[
G_{\ell,i}
=
\nabla_{W_\ell}
\mathcal L
\bigl(f_{\theta_0}(x_i),y_i\bigr).
\]
The corresponding output- and input-side second-moment matrices are
\begin{equation}
C_\ell^{\mathrm{out}}
=
\sum\nolimits_i G_{\ell,i}G_{\ell,i}^{\top},
\qquad
C_\ell^{\mathrm{in}}
=
\sum\nolimits_i G_{\ell,i}^{\top}G_{\ell,i}.
\end{equation}

\textbf{Randomized Nystr\"om subspace estimation.} Let $P_{\mathcal S}$ denote the orthogonal projector onto subspace $\mathcal S$. For each side $q\in\{\mathrm{out},\mathrm{in}\}$, we seek the rank-$r$ subspace that captures the largest amount of historical forecast-loss sensitivity:
\begin{equation}
\mathcal S_{\ell,r}^q
=
\underset{\dim(\mathcal S)=r}{\arg\max}\;
\operatorname{tr}\!\left(P_{\mathcal S}C_\ell^q\right).
\label{eq:optimal_sensitivity_subspace}
\end{equation}
By the Rayleigh--Ritz principle~\citep{horn2012matrix}, the solution is spanned by the leading $r$ eigenvectors of $C_\ell^q$. Because these second-moment matrices are too large to construct explicitly, we estimate their dominant eigenspaces using a randomized Nystr\"om sketch.

Specifically, we draw an orthonormalized Gaussian matrix $\Omega_\ell^q\in\mathbb R^{d_\ell^q\times k}$, where $k=r+p$, and form $Y_\ell^q=C_\ell^q\Omega_\ell^q$. The sketches can be accumulated directly from per-sample gradients without materializing $C_\ell^q$:
\begin{equation}
Y_\ell^{\mathrm{out}}
=
\sum_i G_{\ell,i}
\left(G_{\ell,i}^{\top}\Omega_\ell^{\mathrm{out}}\right),
\qquad
Y_\ell^{\mathrm{in}}
=
\sum_i G_{\ell,i}^{\top}
\left(G_{\ell,i}\Omega_\ell^{\mathrm{in}}\right).
\label{eq:nystrom_sketch}
\end{equation}
We then recover the leading eigendirections from the Nystr\"om approximation
\begin{equation}
\widehat C_\ell^q
=
Y_\ell^q
\left[(\Omega_\ell^q)^\top Y_\ell^q\right]^\dagger
(Y_\ell^q)^\top .
\label{eq:nystrom_operator}
\end{equation}
where ${}^\dagger$ denotes the Moore--Penrose pseudoinverse. Its leading $r$ eigenvectors define the output- and input-side sensitivity bases $U_\ell$ and $V_\ell$, respectively. Complete factorization and implementation details are provided in Appendix~\ref{a:subspace}.

Figure~\ref{fig:svd} reveals a pronounced low-dimensional and asymmetric sensitivity structure. The right subspaces are highly concentrated, with the Stage~2 median capturing over $90\%$ of the gradient trace at rank $32$ and about $97\%$ at rank $128$. In contrast, the Stage~1 left subspace captures only about $55\%$ at rank $128$. Forecast-loss sensitivity is therefore substantially more compressible on the input side and becomes increasingly concentrated toward the later decoder stages.

\begin{figure}
    \centering
    \includegraphics[width=1.0\linewidth]{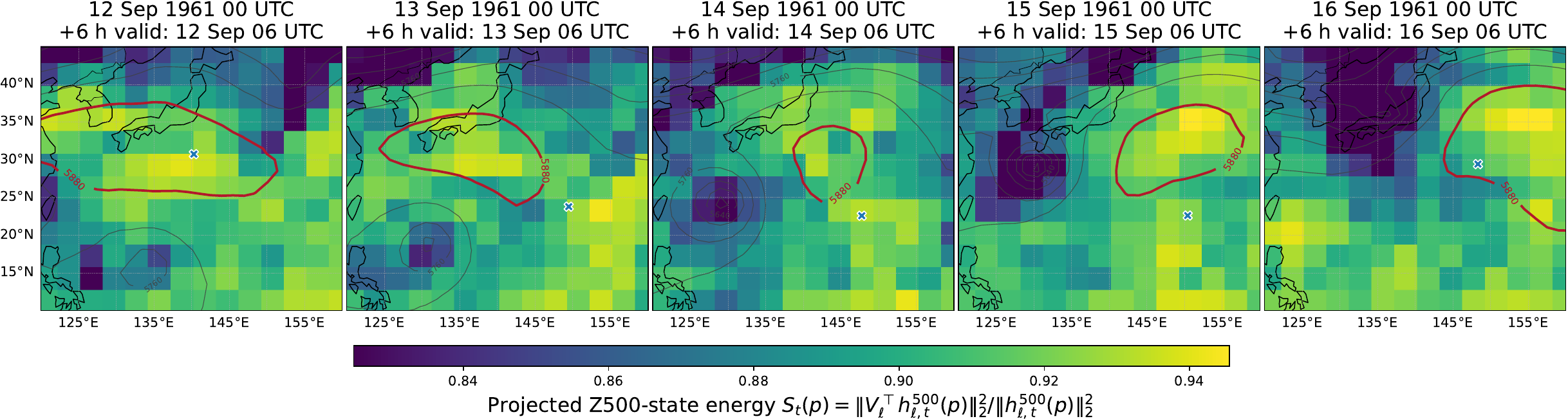}
    \caption{\textbf{Flow-dependent activation of the Z500 forecast-sensitivity subspace.} We visualize five consecutive 00 UTC forecasts from 12--16 September 1961. Colors show the fraction of local Z500 latent-state energy captured by the offline rank-128 sensitivity subspace, $S_t(p)$. Gray contours denote the verifying $500$-hPa geopotential height, the red contour marks the $5880$-gpm boundary conventionally associated with the western North Pacific subtropical high, and the white cross indicates the energy-weighted centroid of the top-$5\%$ sensitivity scores.}
    \label{fig:flow_activation}
\end{figure}

Although the sensitivity bases are estimated offline and remain fixed during deployment, their activation depends on the current atmospheric state. Figure~\ref{fig:flow_activation} shows that regions of high projected Z500-state energy move with the evolving 5880-gpm subtropical-ridge boundary, consistent with NWP studies locating forecast sensitivity around tropical cyclones, their environmental steering flows, and weaknesses in the subtropical ridge~\citep{peng2006sensitivity,reynolds2009recurving,torn2025application}. This state-dependent activation allows a fixed low-dimensional basis to support flow-responsive online corrections.

\subsection{Flow-Dependent Adaptation with Persistent Residual Gating}
\label{sec:residual_adaptation}

With the sensitivity bases established, we translate the operational error structure in Eq.~\ref{eq:structured_error_decomposition} into two complementary online adaptation mechanisms. The flow-dependent correction is learned within the fixed sensitivity subspaces, while persistent errors remaining after this correction are tracked through spatial residual memory. The flow-dependent correction is always applied; a variable-wise gate controls only the additional contribution of the persistent residual correction.

\textbf{Flow-dependent correction.} The forecast-sensitivity bases $U_\ell$ and $V_\ell$ estimated during the offline period remain fixed throughout deployment. For a selected linear layer $\ell$, let $h_{\ell,t}$ denote its current input activation and $W_\ell^0$ its frozen pretrained weight. We introduce an online coefficient matrix $A_{\ell,t}\in\mathbb{R}^{r_\ell^{\mathrm{out}}\times r_\ell^{\mathrm{in}}}$ and modify the layer output as
\begin{equation}
\widetilde a_{\ell,t}
=
W_\ell^0 h_{\ell,t}
+
U_\ell A_{\ell,t}V_\ell^\top h_{\ell,t}.
\label{eq:subspace_activation_correction}
\end{equation}
Here, $V_\ell^\top h_{\ell,t}$ extracts coordinates along historically error-sensitive input directions, $A_{\ell,t}$ maps these coordinates to correction coefficients, and $U_\ell$ maps them into the output-side sensitivity subspace. Because the projected coordinates depend on the current activation $h_{\ell,t}$, the resulting correction remains state-dependent even though the sensitivity bases themselves are fixed. We denote the resulting forecast by
\[
p_{t,c}(x_t;s)
=
f_{\theta_0,c}(x_t;s)
+
d_{t,c}(x_t;s).
\]

\textbf{Persistent residual memory.} To track geographically structured errors that persist after flow-dependent adaptation, we maintain a spatial residual memory $b_{t,c}^{(i)}(s)$ for each atmospheric variable $c$ and initialization-time stream $i\in\{00\,\mathrm{UTC},12\,\mathrm{UTC}\}$. Separate states allow the two forecast cycles to capture systematic differences between initialization times. We parameterize the update rate $\alpha$ by a half-life $H$,
$
\alpha=1-2^{-\Delta t/H},
$
where $\Delta t$ is the interval between successive updates within the same stream. Once the corresponding verification becomes available, the residual memory is updated as
\begin{equation}
r_{t,c}(s)
=
y_{t+1,c}(s)-p_{t,c}(x_t;s),
\qquad
b_{t+1,c}^{(i)}(s)
=
(1-\alpha)b_{t,c}^{(i)}(s)
+
\alpha r_{t,c}(s).
\label{eq:ewma_residual}
\end{equation}
Thus, the memory tracks persistent residual errors remaining after the flow-dependent correction rather than biases of the frozen backbone alone.
\begin{figure}[t]
    \centering
    \includegraphics[width=0.90\linewidth]{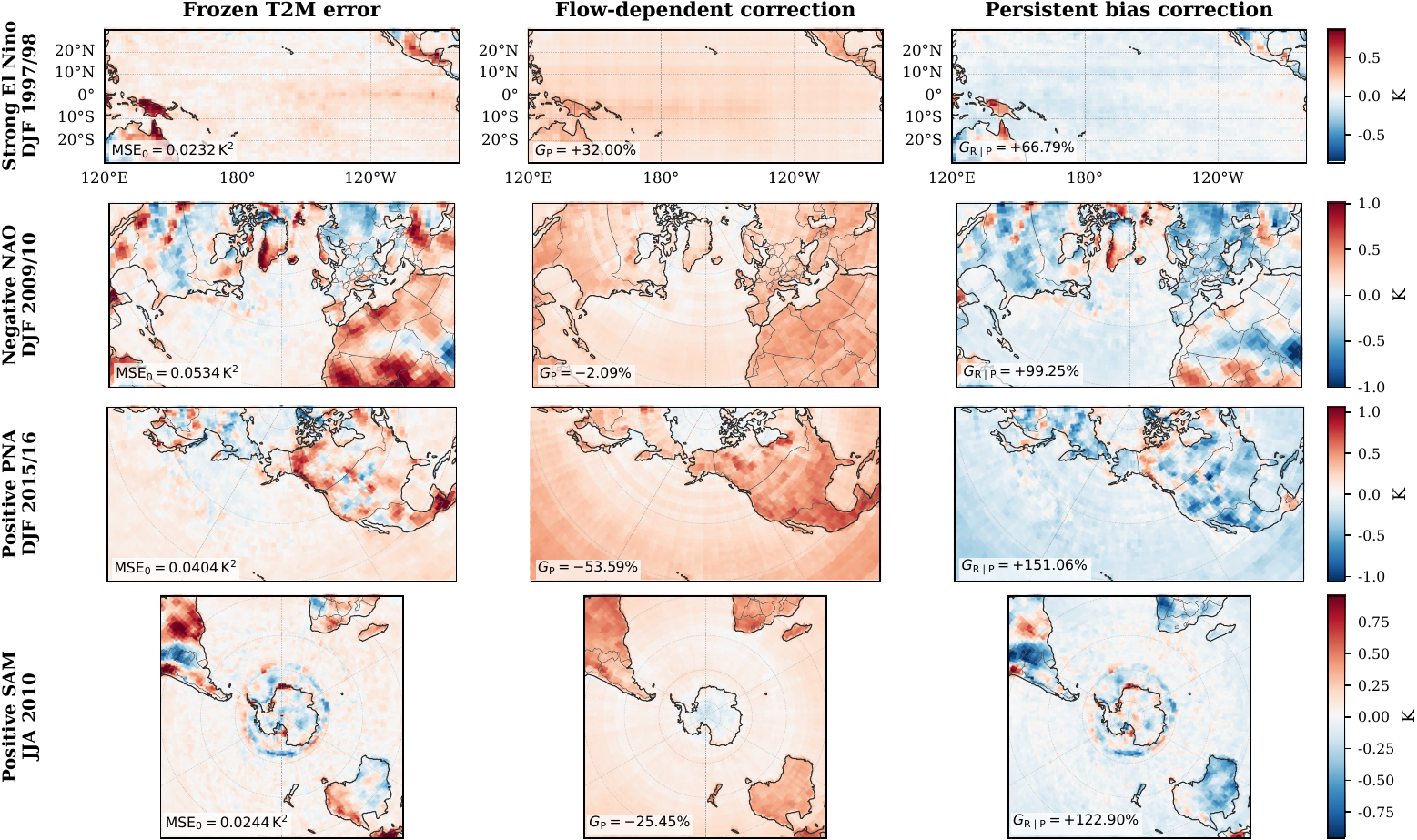}
    \caption{\textbf{Decomposition of T2M forecast-error correction across large-scale circulation regimes.} Rows correspond to a strong El~Ni\~no (DJF 1997/98), negative NAO (DJF 2009/10), positive PNA (DJF 2015/16), and positive SAM (JJA 2010). The frozen-model MSE and the relative contributions of the flow-dependent ($G_{\mathrm{P}}$) and persistent ($G_{\mathrm{R}\mid\mathrm{P}}$) corrections are reported in each panel.}
    \label{fig:climate}
\end{figure}

\textbf{Variable-wise residual gating.} For each atmospheric variable $c$, we maintain a gating weight $q_{t,c}\in[0,1]$ that controls the usefulness of adding the persistent residual correction on top of the flow-corrected forecast:
\begin{equation}
\widehat y_{t+1,c}(s)
=
f_{\theta_0,c}(x_t;s)
+
d_{t,c}(x_t;s)
+
q_{t,c} b_{t,c}^{(i)}(s).
\label{eq:routed_forecast}
\end{equation}
Both the residual memory and the gating weight used at round $t$ depend only on previously revealed verification. After the verifying target $y_{t+1}(s)$ becomes available, we update the flow-dependent adapter coefficients using the forecast loss:
\begin{equation}
A_{\ell,t+1}
=
A_{\ell,t}
-
\eta
\nabla
\mathcal L(\widehat y_{t+1}(s),y_{t+1}(s)),
\label{eq:online_subspace_update}
\end{equation}
where $\eta$ denotes the online learning rate.
After observing $y_{t+1,c}(s)$, we evaluate two endpoint forecasts: the flow-corrected forecast alone and the same forecast with the full persistent residual correction. Their losses are
\begin{equation}
\begin{aligned}
\ell_{t,c,e_1}
&=
\left\|
p_{t,c}(x_t;s)-y_{t+1,c}(s)
\right\|_{2}^{2},\\
\ell_{t,c,e_2}
&=
\left\|
p_{t,c}(x_t;s)+b_{t,c}^{(i)}(s)-y_{t+1,c}(s)
\right\|_{2}^{2}.
\end{aligned}
\label{eq:endpoint_loss}
\end{equation}
Because endpoint losses have strongly field-dependent scales, let $\widetilde\ell_{t,c,e}\in[0,1]$ denote the scale-normalized endpoint loss. We initialize $q_{1,c}=1/2$ and update it by exponential weighting:
\begin{equation}
q_{t+1,c}
=
\frac{
q_{t,c}\exp\!\left(-\eta_q\widetilde{\ell}_{t,c,e_2}\right)
}{
(1-q_{t,c})
\exp\!\left(-\eta_q\widetilde{\ell}_{t,c,e_1}\right)
+
q_{t,c}
\exp\!\left(-\eta_q\widetilde{\ell}_{t,c,e_2}\right)
}.
\label{eq:routing_update}
\end{equation}
Here, $\eta_q>0$ controls how rapidly the gate responds to differences between the two endpoint losses.
\begin{theorem}[Regret against the best fixed correction choice]
\label{thm:gating_regret}
Fix an atmospheric variable $c$. Let $\widetilde{\ell}_{t,c,e_1},\widetilde{\ell}_{t,c,e_2}\in[0,1]$ denote the normalized losses of the flow-corrected forecasts without and with the full persistent residual correction, respectively. Initialize $q_{1,c}=1/2$ and update it by exponential weighting. Then, for any $\eta_q>0$,
\begin{equation}
\sum_{t=1}^{T}
\left[
(1-q_{t,c})\widetilde{\ell}_{t,c,e_1}
+q_{t,c}\widetilde{\ell}_{t,c,e_2}
\right]
-
\min_{e\in\{e_1,e_2\}}
\sum_{t=1}^{T}\widetilde{\ell}_{t,c,e}
\le
\frac{\log 2}{\eta_q}
+\frac{\eta_q T}{8}.
\end{equation}
In particular, choosing $\eta_q=\sqrt{8\log 2/T}$ gives an upper bound of $\sqrt{T\log 2/2}$.
\end{theorem}

Thus, the average regret of the endpoint-weighting objective vanishes as $O(T^{-1/2})$ relative to the better fixed endpoint in hindsight. The guarantee compares flow-dependent correction alone with flow-dependent correction plus the full persistent residual correction, rather than an arbitrary fixed mixture. The proof is provided in Appendix~\ref{A:regret_proof}. We next evaluate whether these complementary adaptation mechanisms improve forecasting skill during long-term deployment.

\section{Experiments}
\begin{table}[t]
\centering
\caption{
Six-hour forecast MSE for eight atmospheric fields.
Original denotes the forecast from the frozen Aurora backbone.
Bold and underlined values indicate the lowest and second-lowest MSE,
respectively. Gain is the percentage reduction in MSE achieved by Ours
relative to the second-best method for each field.
}
\label{tab:selected8_mse_seven_methods}
\setlength{\tabcolsep}{3.5pt}
\renewcommand{\arraystretch}{1.15}
\begin{tabular}{@{}*{10}{c}@{}}
\toprule
Field & Original & $\delta$-Adapter & Proceed & OneNet & FSNet
      & ADAPT-Z & ELF & Ours & Gain (\%) \\
\midrule
Z500 & 1409.729 & \underline{676.367} & 915.514 & 733.233 & 762.317
     & 842.627 & 1409.867 & \textbf{582.499} & 13.88 \\
T850 & 0.12910 & \underline{0.12001} & 0.12290 & 0.12083 & 0.12343
     & 0.12196 & 0.12912 & \textbf{0.11811} & 1.59 \\
U850 & 0.35663 & \underline{0.35459} & 0.35608 & 0.35605 & 0.36020
     & 0.35656 & 0.35667 & \textbf{0.35384} & 0.21 \\
V850 & 0.36702 & \underline{0.36539} & 0.36688 & 0.36686 & 0.37118
     & 0.36732 & 0.36706 & \textbf{0.36485} & 0.15 \\
T2M  & 0.32691 & \underline{0.31019} & 0.32012 & 0.32107 & 0.31981
     & 0.32382 & 0.32694 & \textbf{0.25162} & 18.88 \\
U10  & 0.14060 & \underline{0.13960} & 0.14051 & 0.14050 & 0.14195
     & 0.14060 & 0.14063 & \textbf{0.13853} & 0.77 \\
V10  & 0.14372 & \underline{0.14258} & 0.14367 & 0.14367 & 0.14528
     & 0.14374 & 0.14374 & \textbf{0.14190} & 0.48 \\
MSL  & 1202.503 & \underline{1088.852} & 1166.726 & 1162.141 & 1111.857
     & 1194.673 & 1202.782 & \textbf{976.540} & 10.31 \\
\bottomrule
\end{tabular}
\end{table}
\subsection{Experimental Setup}
\textbf{Dataset and temporal split.}
All experiments use ERA5 reanalysis spanning 1959--2022
(64 years), regridded to a $1.5^\circ$ equiangular grid of
$121\times240$ points. The data are sampled at 6-hour intervals,
corresponding to 00:00 UTC, 06:00 UTC, 12:00 UTC, and 18:00 UTC each day, while forecast
evaluation is performed only for initializations at 00:00 UTC and 12:00 UTC.
Following WeatherBench2~\citep{rasp2024weatherbench}, we use five
pressure-level variables, each defined on 13 pressure levels:
geopotential (Z), specific humidity (Q), temperature (T), and the
zonal and meridional wind components (U and V), together with four
surface variables: 10-m zonal and meridional wind (U10 and V10),
2-m temperature (T2M), and mean sea-level pressure (MSL).
We use the first ten years (1959--1968) exclusively as the offline
pretraining set, while the remaining 54 years (1969--2022) form a
strictly chronological online evaluation stream. Dataset details are
provided in Appendix~\ref{A:Dataset}.

\textbf{Baselines.}
We compare our method with six recent online adaptation approaches:
$\delta$-Adapter~\citep{liang2026deltaadapter},
Proceed~\citep{zhao2025proactive},
OneNet~\citep{zhang2023onenet},
FSNet~\citep{pham2022learning},
ADAPT-Z~\citep{huang2026adaptz}, and
ELF~\citep{elf2025}.
Hyperparameters for all baselines follow the best-performing settings
reported in the corresponding papers.

\textbf{Metrics.}
Following WeatherBench~\citep{rasp2024weatherbench}, we evaluate deterministic forecasts using latitude-area-weighted errors over the global grid. For an atmospheric field \(c\), let \(\hat y_{t,c}(s)\) and \(y_{t,c}(s)\) denote the forecast and ERA5 verification at grid point \(s\) and valid time \(t\), respectively. The latitude weight is $\bar w_s = \frac{\cos\phi_s}{\sum_{s'\in\Omega}\cos\phi_{s'}}$, where \(\phi_s\) is the latitude of grid point \(s\). The two evaluation metrics are defined as follows:
\begin{equation}
\begin{aligned}
\mathrm{MSE}
&=
\frac{1}{|\mathcal T_{\mathrm{test}}|}
\sum_{t\in\mathcal T_{\mathrm{test}}}
\sum_{s\in\Omega}
\bar w_s
\left(\hat y_{t,c}(s)-y_{t,c}(s)\right)^2,\\
\mathrm{MAE}
&=
\frac{1}{|\mathcal T_{\mathrm{test}}|}
\sum_{t\in\mathcal T_{\mathrm{test}}}
\sum_{s\in\Omega}
\bar w_s
\left|\hat y_{t,c}(s)-y_{t,c}(s)\right|.
\end{aligned}
\label{eq:wb2_metrics}
\end{equation}
\textbf{Implementation Details.}
Our default configuration uses an online learning rate of
$2\times10^{-3}$, a sensitivity-subspace rank of 128, and $\eta_q=1$. All experiments are conducted on an
$8\times$NVIDIA A800 GPU cluster. Offline training of Aurora-1.3B
takes approximately 576 GPU-hours, while our online adaptation takes
approximately 57 GPU-hours. Hyperparameter selection and sensitivity
analyses are provided in Appendix~\ref{A:Implementation} and~\ref{A:Hyperparameter Sensitivity}.
\subsection{Online Forecast Performance}
\textbf{Overall comparison.}
We evaluate forecasts at lead times of 6, 12, 24, 48, and 72 hours. Tables~\ref{tab:selected8_mse_seven_methods} and
\ref{tab:selected8_mae_seven_methods} report the 6-hour forecast
performance for eight key atmospheric variables, measured
by mean squared error (MSE) and mean absolute error (MAE).
Averaging the variable-wise relative reductions over these eight
variables, our method reduces MSE and MAE by 5.78\% and 3.08\%,
respectively, compared with the strongest baseline,
$\delta$-Adapter. Averaged over all five lead times, our method
outperforms $\delta$-Adapter by 4.17\% in MSE and 2.76\% in MAE
across the 69 variables. Among the eight key variables,
the largest 6-hour MSE gains over the second-best method occur for
T2M, Z500, and MSL, with reductions of 18.88\%, 13.88\%, and
10.31\%, respectively. Improvements for the wind fields are smaller
but remain consistent, indicating that the proposed adaptation is
particularly effective for thermodynamic and mass-related variables
while retaining gains across the representative fields. Detailed results for all 69 variables at lead times of
6, 12, 24, 48, and 72 hours are provided in
Appendix~\ref{A:numerical Results}.
\begin{table}[t]
\centering
\caption{
Six-hour forecast MAE for eight atmospheric fields.
Original denotes the forecast from the frozen Aurora backbone.
Bold and underlined values indicate the lowest and second-lowest MAE,
respectively. Gain is the percentage reduction in MAE achieved by Ours
relative to the second-best method for each field.
}
\label{tab:selected8_mae_seven_methods}
\setlength{\tabcolsep}{3.5pt}
\renewcommand{\arraystretch}{1.15}
\begin{tabular}{@{}*{10}{c}@{}}
\toprule
Field & Original & $\delta$-Adapter & Proceed & OneNet & FSNet
      & ADAPT-Z & ELF & Ours & Gain (\%) \\
\midrule
Z500 & 29.298 & \underline{20.381} & 23.656 & 21.208 & 21.561
     & 22.565 & 29.298 & \textbf{18.803} & 7.74 \\
T850 & 0.25464 & \underline{0.24135} & 0.24529 & 0.24178 & 0.24544
     & 0.24353 & 0.25464 & \textbf{0.23936} & 0.83 \\
U850 & 0.43165 & \underline{0.43015} & 0.43118 & 0.43114 & 0.43402
     & 0.43155 & 0.43166 & \textbf{0.42938} & 0.18 \\
V850 & 0.43633 & \underline{0.43527} & 0.43626 & 0.43624 & 0.43905
     & 0.43662 & 0.43635 & \textbf{0.43454} & 0.17 \\
T2M  & 0.33799 & \underline{0.33130} & 0.33356 & 0.33415 & 0.33422
     & 0.33541 & 0.33800 & \textbf{0.30100} & 9.15 \\
U10  & 0.26871 & \underline{0.26774} & 0.26856 & 0.26856 & 0.27020
     & 0.26869 & 0.26873 & \textbf{0.26647} & 0.47 \\
V10  & 0.27357 & \underline{0.27248} & 0.27349 & 0.27351 & 0.27521
     & 0.27360 & 0.27358 & \textbf{0.27147} & 0.37 \\
MSL  & 26.135 & \underline{24.726} & 25.687 & 25.620 & 24.971
     & 25.992 & 26.136 & \textbf{23.308} & 5.73 \\
\bottomrule
\end{tabular}
\end{table}

\begin{figure}[htbp]
    \centering
    \includegraphics[width=0.85\linewidth]{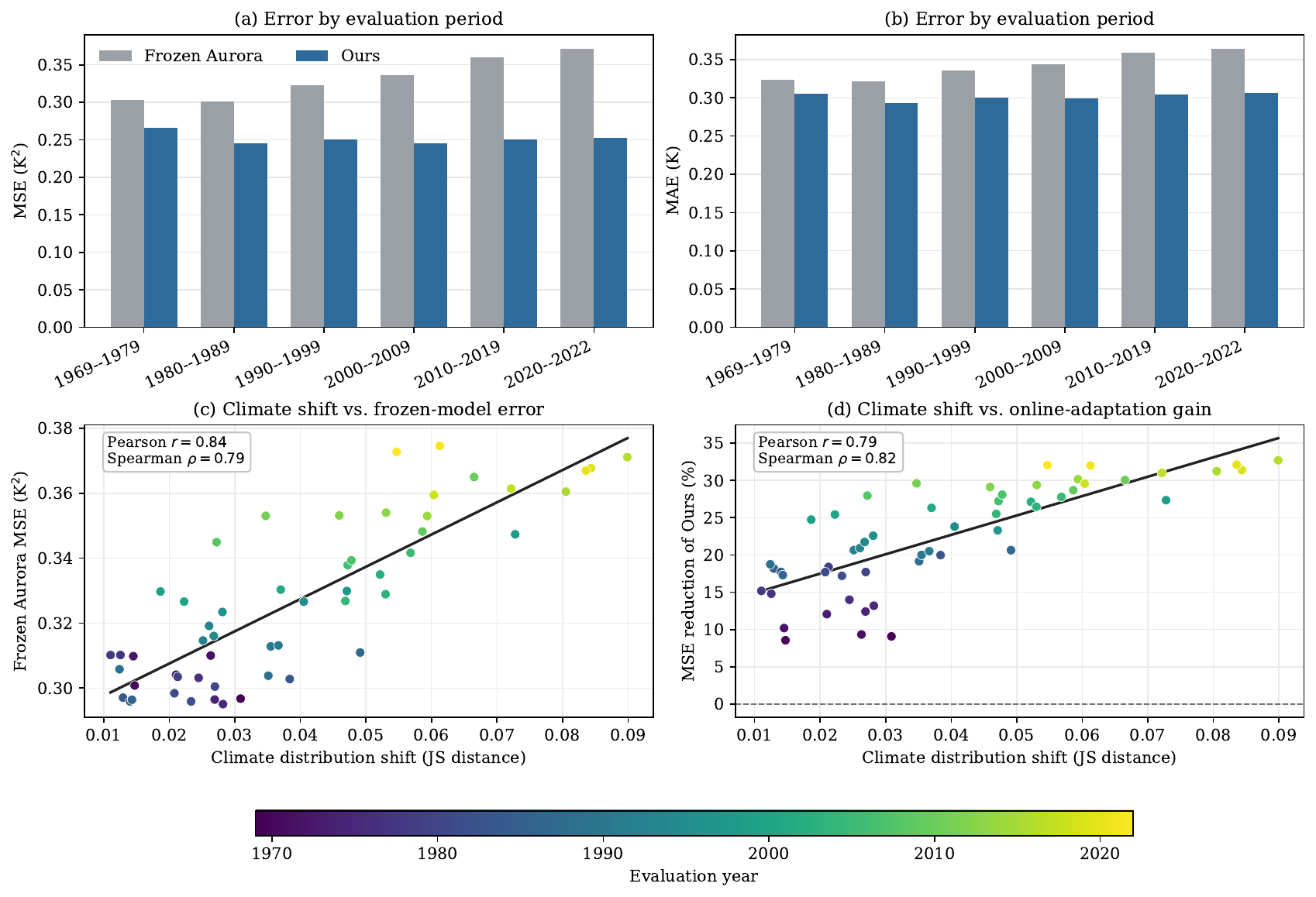}
    \caption{
    T2M forecast error under climate distribution shift.
    Panels (a,b) compare Frozen Aurora and Ours across evaluation
    periods. Panels (c,d) show annual climate shift versus frozen-model
    MSE and the MSE reduction of Ours. Climate shift is measured relative
    to the 1959--1968 distribution using Jensen--Shannon distance.
    }
    \label{fig:t2m_changing_climate}
\end{figure}

\textbf{Temporal drift analysis.}
Figure~\ref{fig:t2m_changing_climate} examines how T2M forecast
performance evolves as the deployment data depart from the
1959--1968 reference period. Across successive evaluation periods,
Frozen Aurora's MSE rises from approximately
$0.30$ to $0.37\,\mathrm{K}^2$, whereas Ours remains comparatively
stable near $0.25\,\mathrm{K}^2$. MAE exhibits a similar pattern,
resulting in a widening performance gap between the frozen and
adapted models.
To quantify distributional change independently of forecast
performance, we compute the annual Jensen--Shannon distance
between ERA5 T2M distributions and the historical reference.
Larger distances are associated with higher frozen-model MSE
(Pearson $r=0.84$) and greater relative MSE reduction from
online adaptation ($r=0.79$). Thus, the benefit of adaptation
is most pronounced in years whose T2M distribution differs more
strongly from the training-period climate. Appendix~\ref{A:variable_wise_climate_shift} extends the
analysis to seven additional atmospheric variables and shows that the
strength of this relationship varies substantially across variables.

\textbf{Gating weight evolution.}
Appendix~\ref{Gating weight} further reveals clear variable-dependent
gating behavior. The persistent-bias endpoint receives dominant weight
($q_t>0.5$) for Z500, T2M, and MSL, with T2M gradually approaching
$q_t=1$, whereas the gate suppresses persistent correction for the wind
fields ($q_t<0.5$). T850 remains close to an equal mixture. The similar
00:00 and 12:00 UTC trajectories indicate that these preferences are robust
across initialization times, supporting the need for variable-specific
adaptive gating. 
\subsection{Ablation and Sensitivity Analysis}
\textbf{Ablation of the two correction components.}
Figure~\ref{fig:endpoint_regret} in Appendix~\ref{A:Ablation of the two correction components} reveals clear variable-dependent roles
of the two corrections. Persistent bias correction substantially improves
Z500, T2M, and MSL, with the strongest effect for T2M, whereas
flow-dependent correction is more effective for T850 and the wind fields.
The case studies further demonstrate that this balance varies with the
atmospheric state. For Z500 during Typhoons Haiyan and Hagibis
(Fig.~\ref{fig:typhoon} in Appendix~\ref{Case-study visualizations}), flow-dependent correction contributes
65.71\% and 30.50\%, respectively, while persistent correction provides
limited or negative additional benefit. In contrast, persistent correction
consistently contributes positively to T2M across four large-scale climate
regimes (Fig.~\ref{fig:climate}), and compensates for unfavorable
flow-dependent corrections in several regimes. These results demonstrate
that the relative importance of persistent and flow-dependent errors varies
across both variables and atmospheric states, motivating their adaptive
combination.

\textbf{Effect of residual-memory timescale.}
Table~\ref{tab:ewma_half_life_sensitivity} in
Appendix~\ref{A:Hyperparameter Sensitivity} examines the sensitivity of our
method to the residual-memory half-life. The 7-day configuration provides the most robust overall performance, achieving the lowest MSE on 29 fields and the lowest MAE on 27 fields. Shortening the half-life to one day makes the residual estimate overly sensitive to transient forecast errors, whereas longer half-lives delay its response to evolving biases. Accordingly, the 1, 30, and 90 day configurations all exhibit lower macro-averaged skill than the 7-day setting, supporting an intermediate residual-memory timescale.

\textbf{Effect of online gating.}
Figure~\ref{fig:endpoint_regret} in Appendix~\ref{A:Ablation of the two correction components} shows that the online gate consistently
improves over both fixed correction strategies by adapting the
contribution of persistent bias correction over time. For Z500, T2M,
and MSL, it largely retains the beneficial persistent component, whereas
for T850 and the wind fields it suppresses this component when it becomes
detrimental. The gated prediction therefore achieves lower cumulative
regret across all variables, indicating that the optimal balance between
persistent and flow-dependent corrections is both variable- and
time-dependent.

\textbf{Comparison with LoRA and online bias correction.}
Tables~\ref{tab:lora_vs_ours_selected8_6h}
and~\ref{tab:ewma_vs_ours_selected8_6h} in
Appendix~\ref{app:lora_bias_correction} compare our method with
rank-128 LoRA and frozen Aurora with online bias correction,
respectively. Our method achieves lower
MSE and MAE than both baselines on all eight selected fields.
The largest relative MSE reduction occurs for Z500, with gains of
5.44\% over LoRA and 6.52\% over online bias correction. These comparisons support the effectiveness of flow-dependent adaptation
and variable-wise gating when used together to correct forecasts.
\section{Conclusion}
We introduced an online adaptation framework for AI weather foundation models under evolving climate distributions. It combines low-dimensional sensitivity updates for flow-dependent errors, residual memory for persistent biases, and a variable-wise gate. On a 54-year chronological Aurora-1.3B evaluation, it consistently outperforms recent online adaptation baselines, with larger climate shifts associated with higher frozen-model errors and greater adaptation gains. These results demonstrate the value of weather-specific error structure for sustaining forecast skill under distribution shift.

\bibliographystyle{plainnat}
\bibliography{refs}

\clearpage
\appendix
\begingroup
  \renewcommand{\undertitle}{\Large Appendix}
  \makeatletter
  \@maketitle
  \makeatother
\endgroup

\counterwithin{equation}{section}
\counterwithin{figure}{section}
\counterwithin{table}{section}
\section{Proof of the online regret bound}
\label{A:regret_proof}

\begin{proof}
Fix an atmospheric variable $c$ and suppress the index $c$
throughout the proof. Write
$\ell_{t,j}:=\widetilde{\ell}_{t,c,e_j}\in[0,1]$
for $j\in\{1,2\}$, and define
$
\pi_{t,1}=1-q_{t,c},
\pi_{t,2}=q_{t,c}.
$
The weighted loss incurred at round $t$ is
\[
m_t
:=
\sum_{j=1}^{2}\pi_{t,j}\ell_{t,j}
=
(1-q_{t,c})\ell_{t,1}
+
q_{t,c}\ell_{t,2}.
\]
Let $\eta:=\eta_q>0$. Introduce unnormalized weights initialized as
$w_{1,1}=w_{1,2}=1/2$ and updated according to
$
w_{t+1,j}
=
w_{t,j}\exp(-\eta\ell_{t,j}), j\in\{1,2\}.
$
Define the potential
$
Z_t:=w_{t,1}+w_{t,2},Z_1=1.
$
Normalizing the weights gives
$\pi_{t,j}=w_{t,j}/Z_t$. In particular,
\[
\pi_{t+1,2}
=
\frac{
\pi_{t,2}\exp(-\eta\ell_{t,2})
}{
\pi_{t,1}\exp(-\eta\ell_{t,1})
+
\pi_{t,2}\exp(-\eta\ell_{t,2})
},
\]
which is precisely the stated exponential-weighting update
for $q_{t+1,c}$.
For the cumulative endpoint loss
\[
L_{t,j}:=\sum_{s=1}^{t}\ell_{s,j},
\qquad L_{0,j}:=0,
\]
the weight recursion implies
\[
w_{t+1,j}
=
\frac{1}{2}\exp(-\eta L_{t,j}).
\]

At each round,
\[
\frac{Z_{t+1}}{Z_t}
=
\sum_{j=1}^{2}
\pi_{t,j}\exp(-\eta\ell_{t,j}).
\]
To bound this quantity, condition on the realized history
and the current endpoint losses, and introduce an auxiliary
random variable $X_t$ satisfying
$
\Pr(X_t=\ell_{t,j})=\pi_{t,j}.
$
Then $X_t\in[0,1]$ and $\mathbb{E}[X_t]=m_t$.
Hoeffding's lemma states that, for any random variable
$X\in[a,b]$ and any $\lambda\in\mathbb{R}$,
\[
\log\mathbb{E}
\left[\exp\bigl(\lambda(X-\mathbb{E}[X])\bigr)\right]
\le
\frac{\lambda^2(b-a)^2}{8}.
\]
Applying this inequality with $\lambda=-\eta$,
$a=0$, and $b=1$ yields
\[
\begin{aligned}
\log\frac{Z_{t+1}}{Z_t}
=
\log\mathbb{E}\left[\exp(-\eta X_t)\right]
=
-\eta m_t
+
\log\mathbb{E}
\left[\exp\bigl(-\eta(X_t-m_t)\bigr)\right]
\le
-\eta m_t+\frac{\eta^2}{8}.
\end{aligned}
\]
Summing over $t=1,\ldots,T$ telescopes the
log-potential ratios:
\[
\log\frac{Z_{T+1}}{Z_1}
\le
-\eta\sum_{t=1}^{T}m_t
+
\frac{\eta^2T}{8}.
\]
Since $Z_1=1$, we obtain
\begin{equation}
\log Z_{T+1}
\le
-\eta\sum_{t=1}^{T}m_t
+
\frac{\eta^2T}{8}.
\label{eq:regret_potential_upper}
\end{equation}

Let
$
j^\star\in\arg\min_{j\in\{1,2\}}L_{T,j}.
$
The potential includes the weight of this endpoint, so
\[
Z_{T+1}
=
\frac{1}{2}\sum_{j=1}^{2}\exp(-\eta L_{T,j})
\ge
\frac{1}{2}\exp(-\eta L_{T,j^\star}).
\]
Taking logarithms gives
\begin{equation}
\log Z_{T+1}
\ge
-\log 2
-
\eta\min_{j\in\{1,2\}}L_{T,j}.
\label{eq:regret_potential_lower}
\end{equation}

Combining
Eqs.~\eqref{eq:regret_potential_upper}
and~\eqref{eq:regret_potential_lower} gives
\[
-\log 2
-
\eta\min_{j\in\{1,2\}}L_{T,j}
\le
-\eta\sum_{t=1}^{T}m_t
+
\frac{\eta^2T}{8}.
\]
Rearranging and dividing by $\eta>0$ yields
\[
\sum_{t=1}^{T}m_t
-
\min_{j\in\{1,2\}}L_{T,j}
\le
\frac{\log 2}{\eta}
+
\frac{\eta T}{8}.
\]
Restoring the original notation proves
\[
\begin{aligned}
\sum_{t=1}^{T}
\left[
(1-q_{t,c})\widetilde{\ell}_{t,c,e_1}
+
q_{t,c}\widetilde{\ell}_{t,c,e_2}
\right]
-
\min_{e\in\{e_1,e_2\}}
\sum_{t=1}^{T}\widetilde{\ell}_{t,c,e}
\le
\frac{\log 2}{\eta_q}
+
\frac{\eta_qT}{8}.
\end{aligned}
\]

The upper bound
\[
B(\eta)=\frac{\log 2}{\eta}+\frac{\eta T}{8}
\]
is minimized over $\eta>0$ when
\[
B'(\eta)
=
-\frac{\log 2}{\eta^2}+\frac{T}{8}
=
0,
\]
which gives
\[
\eta^\star=\sqrt{\frac{8\log 2}{T}}.
\]
Substituting this value yields
\[
B(\eta^\star)
=
\sqrt{\frac{T\log 2}{8}}
+
\sqrt{\frac{T\log 2}{8}}
=
\sqrt{\frac{T\log 2}{2}}.
\]
This completes the proof.
\end{proof}
\section{Dataset}
\label{A:Dataset}
In this section, we are going to introduce the dataset we used in this study detailedly. For the global forecasting, we conduct experiments on the WeatherBench2~\citep{rasp2024weatherbench} benchmark, a subset of ERA5 reanalysis data. The WeatherBench2 benchmark we used is the version of $1.5^\circ$ resolution ($121\times240$), which spans from 1959 to 2022. This subset contains 5 variables (Z, Q, T, U, V) with 13 pressure levels (50 hPa, 100 hPa, 150 hPa, 200 hPa, 250 hPa, 300 hPa, 400 hPa, 500 hPa, 600 hPa, 700 hPa, 850 hPa, 925 hPa and 1,000 hPa) and 4 variables (U10M, V10M, T2M, MSLP) with surface level. All the data we used are shown in Table~\ref{ec_variable}. 
\begin{table*}[htbp]
\centering
\caption{ECMWF variables we used in datasets. The ``Type'' column indicates whether the variable represents a static property, a time-varying single-level property (e.g., surface variables are included), or a time-varying atmospheric property. The ``Long name'' and ``Short name'' columns are ECMWF's labels. The ``level'' column represents whether the variable is a surface variable or an upper-air variable, and indicates how many pressure levels were selected for the upper-air variables. The ``Role'' column indicates whether the variable is something Aurora takes as input and predicts, or only uses as input context.}
\begin{tabular}{lcccc}
\hline
\multicolumn{1}{c}{Type} & Long name                   & Short name & Levels       & Role            \\ \hline
Atmospheric              & Geopotential                & Z          & 13 levels    & Input/Predicted \\
Atmospheric              & Temperature                 & T          & 13 levels    & Input/Predicted \\
Atmospheric              & Specific humidity          & Q          & 13 levels    & Input/Predicted \\
Atmospheric              & U component of wind      & U          & 13 levels    & Input/Predicted \\
Atmospheric              & V component of wind      & V          & 13 levels    & Input/Predicted \\
Single                   & 2m temperature             & T2M        & Single level  & Input/Predicted \\
Single                   & 10m u component of wind & U10        & Single level  & Input/Predicted \\
Single                   & 10m v component of wind & V10        & Single level  & Input/Predicted \\
Single                   & Mean sea level pressure  & MSL        & Single level  & Input/Predicted \\ \hline
Static                   & Land binary mask          & LSM        & Single level & Input           \\
Static                   & Soil type                  & SLT        & Single level & Input           \\
Static                   & Orography                   & OR  & Single level & Input           \\ \hline
\end{tabular}
\label{ec_variable}
\end{table*}
\section{Experimental Details}
\subsection{offline pretraining}
\paragraph{Forecasting Model.}
We use Aurora-1.3B as the forecasting backbone. The model takes two
consecutive atmospheric states as input and predicts the next 6-h state.
It operates on ERA5 fields regridded to a $1.5^\circ$ grid, using five
upper-air variables (Z, Q, T, U, V) on 13 pressure levels together
with four surface variables (T2M, U10, V10, MSL). The backbone uses
an embedding dimension of 512, a patch size of 2, encoder depths
$(6,10,8)$ and decoder depths $(8,10,6)$, with attention heads
$(8,16,32)$ across the encoder stages. Local attention uses a
three-dimensional window of $(2,6,12)$. The Perceiver components use
16 attention heads and four latent levels. Both attention dropout and
drop-path are set to zero. The first ten years of ERA5 (1959--1968)
are used exclusively for offline pretraining, while forecasts are
initialized at 00 and 12 UTC and verified at the corresponding +6-h
valid times.
\paragraph{Training Objective and Hyperparameters.}
The forecasting model is trained using a latitude-weighted multivariate
$\ell_1$ objective over all upper-air and surface variables. The
upper-air variables (Z, Q, T, U, V) are assigned weights
$(3.00,0.60,1.50,0.77,0.54)$, while the surface variables
(T2M, U10, V10, MSL) use
$(3.00,0.77,0.66,1.50)$, respectively. The aggregated upper-air and
surface losses are further weighted by $1.0$ and $0.25$.

We employ a hybrid Muon--AdamW optimizer: Muon is applied to trainable
two-dimensional linear weight matrices, while AdamW is used for the
remaining parameters. Both optimizers use an initial learning rate of
$3\times10^{-4}$ and weight decay of $5\times10^{-6}$.
Muon uses momentum $0.95$ with Nesterov acceleration and five
Newton--Schulz iterations, while AdamW uses
$\beta=(0.9,0.95)$ and $\epsilon=10^{-8}$.
The learning rate follows a cosine decay schedule after an optional
warm-up stage. Training is performed with distributed data parallelism
on eight GPUs, with a per-GPU batch size of 1.
\subsection{Forecast-Sensitivity Subspace Estimation}
\label{a:subspace}
\paragraph{Gradient Collection.}
We estimate forecast-sensitivity directions from the offline period
1959--1968 using forecasts initialized at 00 and 12 UTC with a
6-h lead time. The pretrained Aurora backbone is kept fixed, and
gradients are enabled only temporarily for the two-dimensional
weights of all linear layers in the final two decoder stages and the
surface and atmospheric output heads. For each selected layer $\ell$
and offline sample $i$, we compute
\begin{equation}
    G_{\ell,i}
    =
    \nabla_{W_\ell}
    \mathcal{L}_{\mathrm{off}}
    \bigl(f_{\theta_0}(x_i),y_i\bigr),
\end{equation}
where $\mathcal{L}_{\mathrm{off}}$ is the same normalized,
latitude-weighted multivariate $\ell_1$ loss used for forecasting.
Gradients are computed independently for each sample with batch size
one and are used only to estimate the sensitivity statistics; no model
parameter is updated during this stage. By default, every eighth
offline sample is retained for gradient estimation.
\paragraph{Subspace Estimation and Hyperparameters.}
For each selected weight matrix
$W_\ell\in\mathbb{R}^{d_\ell^{\mathrm{out}}\times
d_\ell^{\mathrm{in}}}$, the per-sample gradient
$G_{\ell,i}$ has the same shape as $W_\ell$. We characterize its
output- and input-side forecast-loss sensitivities using the gradient
second-moment matrices
\begin{equation}
    C_\ell^{\mathrm{out}}
    =
    \sum_i G_{\ell,i}G_{\ell,i}^{\top},
    \qquad
    C_\ell^{\mathrm{in}}
    =
    \sum_i G_{\ell,i}^{\top}G_{\ell,i}.
    \label{eq:appendix_gradient_moments}
\end{equation}
Here,
$C_\ell^{\mathrm{out}}\in
\mathbb{R}^{d_\ell^{\mathrm{out}}\times d_\ell^{\mathrm{out}}}$
and
$C_\ell^{\mathrm{in}}\in
\mathbb{R}^{d_\ell^{\mathrm{in}}\times d_\ell^{\mathrm{in}}}$
describe sensitivity along the output and input dimensions of the layer,
respectively.

Because these matrices are too large to construct explicitly, we estimate
their dominant eigenspaces using a one-pass randomized Nystr\"om sketch.
For each side $q\in\{\mathrm{out},\mathrm{in}\}$, we draw a Gaussian
probe matrix
$\Omega_\ell^q\in\mathbb{R}^{d_\ell^q\times k}$ and orthonormalize its
columns, where $k=r_{\max}+p$. We then form
\begin{equation}
    Y_\ell^q=C_\ell^q\Omega_\ell^q.
    \label{eq:appendix_nystrom_range}
\end{equation}
Importantly, the sketches can be accumulated directly as each per-sample
gradient is observed:
\begin{equation}
    Y_\ell^{\mathrm{out}}
    =
    \sum_i
    G_{\ell,i}
    \left(
        G_{\ell,i}^{\top}\Omega_\ell^{\mathrm{out}}
    \right),
    \qquad
    Y_\ell^{\mathrm{in}}
    =
    \sum_i
    G_{\ell,i}^{\top}
    \left(
        G_{\ell,i}\Omega_\ell^{\mathrm{in}}
    \right).
    \label{eq:appendix_nystrom_accumulation}
\end{equation}
Thus, neither gradient second-moment matrix needs to be materialized or
stored.

After the one-pass accumulation, we construct the small core matrix
\begin{equation}
    B_\ell^q
    =
    (\Omega_\ell^q)^\top Y_\ell^q
    =
    (\Omega_\ell^q)^\top
    C_\ell^q
    \Omega_\ell^q.
    \label{eq:appendix_nystrom_core}
\end{equation}
For numerical stability, we symmetrize the core matrix as
$B_\ell^q\leftarrow
\frac{1}{2}\left(B_\ell^q+(B_\ell^q)^\top\right)$ before computing its
eigendecomposition
\begin{equation}
    B_\ell^q
    =
    S_\ell^q\Lambda_\ell^q(S_\ell^q)^\top.
\end{equation}
Eigenvalues below the relative pseudoinverse tolerance
$\tau_{\mathrm{pinv}}\lambda_{\max}(B_\ell^q)$ are discarded. Let
$S_{\ell,+}^q$ and $\Lambda_{\ell,+}^q$ denote the retained eigenvectors
and positive eigenvalues. The Moore--Penrose pseudoinverse is then
computed as
\begin{equation}
    (B_\ell^q)^\dagger
    =
    S_{\ell,+}^q
    (\Lambda_{\ell,+}^q)^{-1}
    (S_{\ell,+}^q)^\top.
\end{equation}

The corresponding Nystr\"om approximation is
\begin{equation}
    \widehat C_\ell^q
    =
    Y_\ell^q
    (B_\ell^q)^\dagger
    (Y_\ell^q)^\top.
    \label{eq:appendix_nystrom_operator}
\end{equation}
To recover its dominant eigendirections without explicitly constructing
$\widehat C_\ell^q$, we form
\begin{equation}
    F_\ell^q
    =
    Y_\ell^q
    S_{\ell,+}^q
    (\Lambda_{\ell,+}^q)^{-1/2},
\end{equation}
which satisfies
$\widehat C_\ell^q=F_\ell^q(F_\ell^q)^\top$. A compact singular-value
decomposition
\begin{equation}
    F_\ell^q
    =
    \widetilde U_\ell^q
    \Sigma_\ell^q
    (Q_\ell^q)^\top
\end{equation}
therefore gives
\begin{equation}
    \widehat C_\ell^q
    =
    \widetilde U_\ell^q
    (\Sigma_\ell^q)^2
    (\widetilde U_\ell^q)^\top.
\end{equation}
The leading columns of $\widetilde U_\ell^q$ approximate the dominant
eigenvectors of $C_\ell^q$. We consequently define the output- and
input-side sensitivity bases as
\begin{equation}
    U_\ell
    =
    \widetilde U_\ell^{\mathrm{out}}[:,1:r],
    \qquad
    V_\ell
    =
    \widetilde U_\ell^{\mathrm{in}}[:,1:r].
    \label{eq:appendix_sensitivity_bases}
\end{equation}

We estimate a single maximum-rank basis with
$r_{\max}=128$ and use
$r\in\{4,8,16,32,64,128\}$ as nested prefixes of this basis. The
Nystr\"om oversampling dimension is $p=16$, giving a sketch size of
$k=144$, and the random seed is fixed to $2026$. Unless otherwise
specified, all experiments use $r=128$. Once estimated from the offline
period, $U_\ell$ and $V_\ell$ remain fixed throughout online adaptation;
only the projected adapter coefficients are updated.
\subsection{Online Learning}
\paragraph{Hyperparameter}
During online adaptation, all pretrained Aurora parameters remain frozen
and only the low-dimensional projected coefficients are updated. We use
SGD with a learning rate of $2\times10^{-3}$, zero momentum, and no
weight decay. The forecast-sensitivity adapters are applied to the final
two decoder stages and output heads with rank $r=128$. The adaptive gate uses $\eta_q=1.0$, and a maximum normalized loss of 1.
Online adaptation proceeds sequentially with batch size one over
1969--2022, with separate residual and routing states for 00 and 12 UTC.
All updates are performed only after the corresponding 6-h verification
becomes available, ensuring a strictly causal evaluation protocol.
\label{A:Implementation}
\subsection{Hyperparameter Sensitivity}
\label{A:Hyperparameter Sensitivity}
\begin{table}[htbp]
\centering
\small
\caption{Sensitivity to the forecast-sensitivity subspace rank,
with $\eta_A=2\times10^{-3}$ and $\eta_q=1$ fixed. Changes are
relative to the default rank of 128.}
\label{tab:subspace_rank_sensitivity}
\setlength{\tabcolsep}{7pt}
\begin{tabular}{lrrrr}
\toprule
Rank & Average MSE & $\Delta$MSE (\%) &
Average MAE & $\Delta$MAE (\%) \\
\midrule
16  & 205.2233 & +1.07 & 5.03718 & +0.52 \\
32  & 204.5598 & +0.75 & 5.02971 & +0.37 \\
64  & 203.8227 & +0.38 & 5.02055 & +0.19 \\
128 (default) & \textbf{203.0428} & --
              & \textbf{5.01111} & -- \\
\bottomrule
\end{tabular}
\end{table}
\begin{table}[htbp]
\centering
\small
\caption{Sensitivity to the online adaptation learning rate
$\eta$, with rank 128 and $\eta_q=1$ fixed. Changes are relative
to the default $\eta=2\times10^{-3}$; positive values indicate
higher error.}
\label{tab:online_lr_sensitivity}
\setlength{\tabcolsep}{7pt}
\begin{tabular}{lrrrr}
\toprule
$\eta_A$ & Average MSE & $\Delta$MSE (\%) &
Average MAE & $\Delta$MAE (\%) \\
\midrule
$5\times10^{-4}$ & 204.4093 & +0.67 & 5.02476 & +0.27 \\
$\mathbf{1\times10^{-3}}$ & \textbf{202.6691} & \textbf{$-0.18$}
                         & \textbf{5.00347} & \textbf{$-0.15$} \\
$2\times10^{-3}$ (default) & 203.0428 & -- & 5.01111 & -- \\
$3\times10^{-3}$ & 227.4027 & +12.00 & 5.30524 & +5.87 \\
$5\times10^{-3}$ & 475.6694 & +134.27 & 7.56256 & +50.92 \\
\bottomrule
\end{tabular}
\end{table}
\textbf{Subspace rank.}
Table~\ref{tab:subspace_rank_sensitivity} shows a monotonic decrease
in average MSE and MAE as the rank increases from 16 to 128.
Rank 128 also improves both metrics for all 69 fields relative
to rank 16. The modest aggregate differences indicate limited
sensitivity to the precise rank.

\textbf{Online learning rate.}
Table~\ref{tab:online_lr_sensitivity} shows that $\eta=10^{-3}$
achieves the lowest average MSE and MAE, improving on the default
$2\times10^{-3}$ by 0.18\% and 0.15\%, respectively. However, the
default performs better on 62 of the 69 fields. Larger learning
rates substantially degrade geopotential-height forecasts,
despite improving several other fields.

\textbf{Gate learning rate.}
Table~\ref{tab:router_eta_sensitivity} shows that the default
$\eta_q=1$ achieves the lowest average MSE and MAE. Both smaller
and larger values increase aggregate error, although the changes
around the default are relatively small.
\begin{table}[htbp]
\centering
\small
\caption{Sensitivity to the router learning rate $\eta_q$,
with rank 128 and $\eta_A=2\times10^{-3}$ fixed. Changes are
relative to the default $\eta_q=1$.}
\label{tab:router_eta_sensitivity}
\setlength{\tabcolsep}{7pt}
\begin{tabular}{lrrrr}
\toprule
$\eta_q$ & Average MSE & $\Delta$MSE (\%) &
Average MAE & $\Delta$MAE (\%) \\
\midrule
0.25 & 205.5877 & +1.25 & 5.04191 & +0.61 \\
0.5  & 204.0712 & +0.51 & 5.02413 & +0.26 \\
1 (default) & \textbf{203.0428} & --
            & \textbf{5.01111} & -- \\
2    & 203.5988 & +0.27 & 5.01742 & +0.13 \\
4    & 203.9941 & +0.47 & 5.02265 & +0.23 \\
\bottomrule
\end{tabular}
\end{table}
\begin{table}[htbp]
\centering
\small
\caption{
Sensitivity to the residual-memory half-life.
$\Delta$MSE and $\Delta$MAE denote the macro-averaged relative
improvement over the 7-day configuration; higher is better.
Best-field counts indicate the number of the 69 atmospheric variables
on which each configuration achieves the lowest error.
}
\label{tab:ewma_half_life_sensitivity}
\begin{tabular}{ccccc}
\toprule
Half-life &
$\Delta$MSE (\%) $\uparrow$ &
$\Delta$MAE (\%) $\uparrow$ &
MSE wins $\uparrow$ &
MAE wins $\uparrow$ \\
\midrule
1 day  & $-0.900$ & $-0.418$ & 4  & 9  \\
7 days & \textbf{0.000} & \textbf{0.000} &
         \textbf{29} & \textbf{27} \\
30 days & \underline{$-0.621$} & \underline{$-0.253$} &
          \underline{22} & \underline{21} \\
90 days & $-1.494$ & $-0.678$ & 14 & 12 \\
\bottomrule
\end{tabular}
\end{table}
\subsection{Comparison with Random Subspaces and LoRA}
\label{app:subspace_lora_comparison}
Table~\ref{tab:lora_random_subspace_all69} compares our
forecast-sensitivity adaptation with random-subspace adaptation
and LoRA at rank 128. Ours achieves the lowest average MSE and
MAE, reducing these errors by 2.11\% and 1.00\% relative to the
parameter-matched random-subspace control, and by 5.58\% and
2.91\% relative to LoRA. The improvement over random subspaces
supports the value of selecting adaptation directions from
historical forecast-loss gradients, beyond the benefit of
restricting updates to a low-dimensional space.
\begin{table}[t]
\centering
\small
\caption{
Comparison of forecast-sensitivity adaptation with random-subspace
adaptation and LoRA at rank 128.
Results report six-hour forecasting errors averaged over 69 atmospheric
variables on 39,440 common forecast initializations from 1969--2022.
Gains are computed as the relative reduction in the averaged errors
achieved by Ours over each baseline.
}
\label{tab:lora_random_subspace_all69}
\setlength{\tabcolsep}{8pt}
\begin{tabular}{lcccc}
\toprule
Method
& Average MSE
& Average MAE
& MSE Gain (\%)
& MAE Gain (\%) \\
\midrule
LoRA
& 215.032
& 5.1614
& 5.58
& 2.91 \\
Random Subspace
& 207.412
& 5.0618
& 2.11
& 1.00 \\
Ours
& \textbf{203.041}
& \textbf{5.0111}
& --
& -- \\
\bottomrule
\end{tabular}
\end{table}
\subsection{Comparisons with LoRA and Bias Correction}
\label{app:lora_bias_correction}
Tables~\ref{tab:lora_vs_ours_selected8_6h} and
\ref{tab:ewma_vs_ours_selected8_6h} report six-hour forecast errors for
eight atmospheric fields. Our method achieves lower MSE and MAE than
both rank-128 LoRA and the frozen model with online Bias Correction on all eight fields. Relative to LoRA, the largest MSE
reduction occurs for Z500 (5.44\%), while reductions for the four wind
fields are below 0.4\%. Relative to bias correction alone, MSE reductions
range from 2.47\% for T2M to 6.52\% for Z500. This comparison measures
the combined benefit of the sensitivity-based update and the variable-wise
gate over residual correction alone.
\begin{table}[t]
\centering
\small
\caption{Six-hour forecast errors for LoRA and Ours on eight atmospheric
fields. Bold indicates the lower error for each field and metric.}
\label{tab:lora_vs_ours_selected8_6h}
\vspace{6pt}
\setlength{\tabcolsep}{10pt}
\renewcommand{\arraystretch}{1.15}
\begin{tabular}{@{}*{5}{c}@{}}
\toprule
& \multicolumn{2}{c}{MSE} & \multicolumn{2}{c}{MAE} \\
\cmidrule(lr){2-3}\cmidrule(lr){4-5}
Field & LoRA & Ours & LoRA & Ours \\
\midrule
Z500 & 615.978 & \textbf{582.498} & 19.385  & \textbf{18.803}  \\
T850 & 0.11914 & \textbf{0.11811} & 0.24111 & \textbf{0.23936} \\
U850 & 0.35417 & \textbf{0.35384} & 0.42970 & \textbf{0.42938} \\
V850 & 0.36512 & \textbf{0.36485} & 0.43482 & \textbf{0.43454} \\
T2M  & 0.25601 & \textbf{0.25162} & 0.30298 & \textbf{0.30099} \\
U10  & 0.13895 & \textbf{0.13853} & 0.26690 & \textbf{0.26647} \\
V10  & 0.14231 & \textbf{0.14190} & 0.27189 & \textbf{0.27147} \\
MSL  & 997.474 & \textbf{976.539} & 23.567  & \textbf{23.308}  \\
\bottomrule
\end{tabular}
\end{table}
\begin{table}[t]
\centering
\small
\caption{Six-hour forecast errors for bias correction and Ours on eight
atmospheric fields. Bold indicates the lower error for each field and metric.}
\label{tab:ewma_vs_ours_selected8_6h}
\vspace{6pt}
\setlength{\tabcolsep}{7pt}
\renewcommand{\arraystretch}{1.15}
\begin{tabular}{@{}*{5}{c}@{}}
\toprule
& \multicolumn{2}{c}{MSE} & \multicolumn{2}{c}{MAE} \\
\cmidrule(lr){2-3}\cmidrule(lr){4-5}
Field & Frozen + online bias correction & Ours & Frozen + online bias correction & Ours \\
\midrule
Z500 & 623.095  & \textbf{582.499} & 19.494  & \textbf{18.803}  \\
T850 & 0.12173  & \textbf{0.11811} & 0.24476 & \textbf{0.23936} \\
U850 & 0.36613  & \textbf{0.35384} & 0.43846 & \textbf{0.42938} \\
V850 & 0.37764  & \textbf{0.36485} & 0.44379 & \textbf{0.43454} \\
T2M  & 0.25800  & \textbf{0.25162} & 0.30453 & \textbf{0.30100} \\
U10  & 0.14336  & \textbf{0.13853} & 0.27212 & \textbf{0.26647} \\
V10  & 0.14696  & \textbf{0.14190} & 0.27725 & \textbf{0.27147} \\
MSL  & 1015.803 & \textbf{976.540} & 23.778  & \textbf{23.308}  \\
\bottomrule
\end{tabular}
\end{table}
\clearpage
\subsection{Computational Efficiency Analysis}
\begin{table}[htbp]
\centering
\small
\caption{
Computational efficiency over the complete 1969--2022
evaluation stream. Runtime denotes total wall-clock time,
GPU memory denotes the observed device-memory usage, and
theoretical FP32 operations are estimated by multiplying runtime
by the 19.5-TFLOP/s peak FP32 throughput of one NVIDIA A800 GPU.
Bold and underline indicate the lowest and second-lowest values.
}
\label{tab:efficiency_comparison}
\setlength{\tabcolsep}{7pt}
\begin{tabular}{lccc}
\toprule
Method
& Runtime (h)
& GPU Memory (GiB)
& FP32 Ops. (EFLOPs) \\
\midrule
$\delta$-Adapter
& 57.43
& 12.71
& 4.032 \\

Proceed
& 56.61
& 22.98
& 3.974 \\

OneNet
& \textbf{56.16}
& \textbf{9.43}
& \textbf{3.942} \\

FSNet
& 57.24
& 15.09
& 4.018 \\

Adapt-Z
& 81.08
& 13.70
& 5.692 \\

ELF
& \underline{56.21}
& \underline{9.45}
& \underline{3.946} \\

Ours
& 57.49
& 20.09
& 4.036 \\
\bottomrule
\end{tabular}
\end{table}
We further evaluate the computational efficiency of the proposed method and several mainstream online adaptation baselines on the full 1969--2022 evaluation stream. Table \ref{tab:efficiency_comparison} reports the comparison in terms of total wall-clock runtime, GPU memory consumption, and theoretical FP32 computational overhead.

As shown in the table~\ref{tab:efficiency_comparison}, lightweight adaptation methods including OneNet and ELF achieve the best and second-best resource efficiency. Specifically, OneNet yields the lowest runtime (56.16 h), GPU memory usage (9.43 GiB), and FP32 operations (3.942 EFLOPs), while ELF exhibits nearly identical efficiency with only tiny increments in all metrics. Our method maintains moderate computational cost over the evaluation period. Although it does not attain the minimal resource footprint, it attains a good trade-off between computational efficiency and forecasting accuracy, avoiding the notable performance drop observed in ultra-lightweight baselines. By contrast, Adapt-Z consumes substantially more runtime and computation, showing inferior efficiency for long sequential online weather adaptation.
\clearpage
\section{Detailed Results}
\subsection{Numerical Results}
\label{A:numerical Results}

\begingroup
\scriptsize
\setlength{\tabcolsep}{3pt}
\renewcommand{\arraystretch}{0.95}
\begin{longtable}{lrrrrrrrr}
\caption{MSE for all 69 output fields at a 6-hour lead time. Bold and underline indicate the lowest and second-lowest values, respectively. Gain is the relative MSE reduction of Ours over the best baseline for each field; negative values indicate that Ours has a larger error.}\label{tab:all69_mse_6h_seven_methods} \\
\toprule
Field & $\delta$-Adapter & Proceed & OneNet & FSNet & ADAPT-Z & ELF & Ours & Gain (\%) \\
\midrule
\endfirsthead
\multicolumn{9}{c}{\tablename\ \thetable{} -- continued} \\
\toprule
Field & $\delta$-Adapter & Proceed & OneNet & FSNet & ADAPT-Z & ELF & Ours & Gain (\%) \\
\midrule
\endhead
\midrule
\multicolumn{9}{r}{Continued on next page} \\
\endfoot
\bottomrule
\endlastfoot
Z50 & \underline{4209.465} & 5437.505 & 4607.417 & 4504.681 & 4948.431 & 6066.945 & \textbf{3585.281} & \textbf{14.83} \\
Z100 & \underline{1823.899} & 2982.668 & 2091.184 & 2185.517 & 2518.817 & 4014.980 & \textbf{1547.844} & \textbf{15.14} \\
Z150 & \underline{1377.309} & 2397.649 & 1504.826 & 1583.329 & 1944.765 & 3980.225 & \textbf{1147.644} & \textbf{16.67} \\
Z200 & \underline{1159.681} & 1990.212 & 1231.392 & 1359.458 & 1586.408 & 3593.339 & \textbf{973.499} & \textbf{16.05} \\
Z250 & \underline{1038.302} & 1719.849 & 1102.080 & 1215.049 & 1404.534 & 3116.652 & \textbf{881.763} & \textbf{15.08} \\
Z300 & \underline{953.204} & 1507.493 & 1012.962 & 1131.921 & 1252.109 & 2691.851 & \textbf{815.082} & \textbf{14.49} \\
Z400 & \underline{796.685} & 1161.953 & 859.637 & 904.398 & 1027.819 & 1928.172 & \textbf{686.996} & \textbf{13.77} \\
Z500 & \underline{676.367} & 915.514 & 733.233 & 762.317 & 842.627 & 1409.867 & \textbf{582.499} & \textbf{13.88} \\
\end{longtable}
\endgroup
\clearpage

\begingroup
\scriptsize
\setlength{\tabcolsep}{3pt}
\renewcommand{\arraystretch}{0.95}
\begin{longtable}{lrrrrrrrr}
\caption{MAE for all 69 output fields at a 6-hour lead time. Bold and underline indicate the lowest and second-lowest values, respectively. Gain is the relative MAE reduction of Ours over the best baseline for each field; negative values indicate that Ours has a larger error.}\label{tab:all69_mae_6h_seven_methods} \\
\toprule
Field & $\delta$-Adapter & Proceed & OneNet & FSNet & ADAPT-Z & ELF & Ours & Gain (\%) \\
\midrule
\endfirsthead
\multicolumn{9}{c}{\tablename\ \thetable{} -- continued} \\
\toprule
Field & $\delta$-Adapter & Proceed & OneNet & FSNet & ADAPT-Z & ELF & Ours & Gain (\%) \\
\midrule
\endhead
\midrule
\multicolumn{9}{r}{Continued on next page} \\
\endfoot
\bottomrule
\endlastfoot
Z50 & \underline{50.485} & 57.351 & 52.790 & 52.133 & 54.616 & 60.939 & \textbf{46.383} & \textbf{8.12} \\
Z100 & \underline{33.516} & 42.085 & 35.773 & 36.259 & 38.567 & 48.570 & \textbf{30.720} & \textbf{8.34} \\
Z150 & \underline{29.126} & 37.653 & 30.315 & 30.900 & 33.752 & 48.033 & \textbf{26.405} & \textbf{9.34} \\
Z200 & \underline{26.711} & 34.440 & 27.462 & 28.618 & 30.560 & 45.727 & \textbf{24.292} & \textbf{9.06} \\
Z250 & \underline{25.252} & 32.101 & 25.966 & 27.066 & 28.789 & 42.761 & \textbf{23.096} & \textbf{8.54} \\
Z300 & \underline{24.179} & 30.128 & 24.873 & 26.134 & 27.246 & 39.958 & \textbf{22.186} & \textbf{8.24} \\
Z400 & \underline{22.092} & 26.565 & 22.920 & 23.409 & 24.789 & 34.087 & \textbf{20.370} & \textbf{7.79} \\
Z500 & \underline{20.381} & 23.656 & 21.208 & 21.561 & 22.565 & 29.298 & \textbf{18.803} & \textbf{7.74} \\
\end{longtable}
\endgroup
\clearpage

\begingroup
\scriptsize
\setlength{\tabcolsep}{3pt}
\renewcommand{\arraystretch}{0.95}
\begin{longtable}{lrrrrrrrr}
\caption{MSE for all 69 output fields at a 12-hour lead time. Bold and underline indicate the lowest and second-lowest values, respectively. Gain is the relative MSE reduction of Ours over the best baseline for each field; negative values indicate that Ours has a larger error.}\label{tab:all69_mse_12h_seven_methods} \\
\toprule
Field & $\delta$-Adapter & Proceed & OneNet & FSNet & ADAPT-Z & ELF & Ours & Gain (\%) \\
\midrule
\endfirsthead
\multicolumn{9}{c}{\tablename\ \thetable{} -- continued} \\
\toprule
Field & $\delta$-Adapter & Proceed & OneNet & FSNet & ADAPT-Z & ELF & Ours & Gain (\%) \\
\midrule
\endhead
\midrule
\multicolumn{9}{r}{Continued on next page} \\
\endfoot
\bottomrule
\endlastfoot
Z50 & \underline{14365.431} & 17979.110 & 15225.157 & 15752.742 & 16317.057 & 20452.004 & \textbf{9734.218} & \textbf{32.24} \\
Z100 & \underline{8748.692} & 12367.749 & 9460.087 & 10411.313 & 10883.791 & 15764.031 & \textbf{6154.921} & \textbf{29.65} \\
Z150 & \underline{8577.524} & 12002.753 & 8979.619 & 9983.670 & 10614.123 & 17284.912 & \textbf{6143.984} & \textbf{28.37} \\
Z200 & \underline{8828.636} & 11628.980 & 9082.756 & 10302.973 & 10409.702 & 16520.377 & \textbf{6574.119} & \textbf{25.54} \\
Z250 & \underline{8807.777} & 11028.791 & 9011.525 & 9995.923 & 10113.434 & 15105.278 & \textbf{6926.823} & \textbf{21.36} \\
Z300 & \underline{8136.609} & 9904.915 & 8323.254 & 9205.649 & 9183.314 & 13219.004 & \textbf{6615.908} & \textbf{18.69} \\
Z400 & \underline{5980.176} & 7065.809 & 6150.743 & 6582.951 & 6715.863 & 9097.567 & \textbf{5051.664} & \textbf{15.53} \\
Z500 & \underline{4479.378} & 5146.016 & 4624.836 & 4888.316 & 4965.632 & 6393.002 & \textbf{3876.690} & \textbf{13.45} \\
\end{longtable}
\endgroup
\clearpage

\begingroup
\scriptsize
\setlength{\tabcolsep}{3pt}
\renewcommand{\arraystretch}{0.95}
\begin{longtable}{lrrrrrrrr}
\caption{MAE for all 69 output fields at a 12-hour lead time. Bold and underline indicate the lowest and second-lowest values, respectively. Gain is the relative MAE reduction of Ours over the best baseline for each field; negative values indicate that Ours has a larger error.}\label{tab:all69_mae_12h_seven_methods} \\
\toprule
Field & $\delta$-Adapter & Proceed & OneNet & FSNet & ADAPT-Z & ELF & Ours & Gain (\%) \\
\midrule
\endfirsthead
\multicolumn{9}{c}{\tablename\ \thetable{} -- continued} \\
\toprule
Field & $\delta$-Adapter & Proceed & OneNet & FSNet & ADAPT-Z & ELF & Ours & Gain (\%) \\
\midrule
\endhead
\midrule
\multicolumn{9}{r}{Continued on next page} \\
\endfoot
\bottomrule
\endlastfoot
Z50 & \underline{90.390} & 102.544 & 92.985 & 95.301 & 96.925 & 110.463 & \textbf{73.010} & \textbf{19.23} \\
Z100 & \underline{70.222} & 84.863 & 73.360 & 77.346 & 78.755 & 95.140 & \textbf{58.131} & \textbf{17.22} \\
Z150 & \underline{68.098} & 82.814 & 69.932 & 74.813 & 76.753 & 100.329 & \textbf{56.762} & \textbf{16.65} \\
Z200 & \underline{67.695} & 80.703 & 68.932 & 75.294 & 74.980 & 99.348 & \textbf{57.160} & \textbf{15.56} \\
Z250 & \underline{66.354} & 77.603 & 67.315 & 73.047 & 72.872 & 94.526 & \textbf{57.310} & \textbf{13.63} \\
Z300 & \underline{63.079} & 72.891 & 63.994 & 69.600 & 68.715 & 88.044 & \textbf{55.254} & \textbf{12.41} \\
Z400 & \underline{53.739} & 61.082 & 54.776 & 58.232 & 58.513 & 72.624 & \textbf{48.079} & \textbf{10.53} \\
Z500 & \underline{46.709} & 51.972 & 47.719 & 50.267 & 50.384 & 60.542 & \textbf{42.400} & \textbf{9.22} \\
\end{longtable}
\endgroup
\clearpage

\begingroup
\scriptsize
\setlength{\tabcolsep}{3pt}
\renewcommand{\arraystretch}{0.95}
\begin{longtable}{lrrrrrrrr}
\caption{MSE for all 69 output fields at a 24-hour lead time. Bold and underline indicate the lowest and second-lowest values, respectively. Gain is the relative MSE reduction of Ours over the best baseline for each field; negative values indicate that Ours has a larger error.}\label{tab:all69_mse_24h_seven_methods} \\
\toprule
Field & $\delta$-Adapter & Proceed & OneNet & FSNet & ADAPT-Z & ELF & Ours & Gain (\%) \\
\midrule
\endfirsthead
\multicolumn{9}{c}{\tablename\ \thetable{} -- continued} \\
\toprule
Field & $\delta$-Adapter & Proceed & OneNet & FSNet & ADAPT-Z & ELF & Ours & Gain (\%) \\
\midrule
\endhead
\midrule
\multicolumn{9}{r}{Continued on next page} \\
\endfoot
\bottomrule
\endlastfoot
Z50 & \underline{26384.444} & 35494.019 & 29055.287 & 30083.238 & 31629.178 & 43581.490 & \textbf{20660.919} & \textbf{21.69} \\
Z100 & \underline{16474.766} & 24327.944 & 18396.249 & 19935.711 & 21189.427 & 34017.417 & \textbf{12972.019} & \textbf{21.26} \\
Z150 & \underline{17020.844} & 24122.492 & 18115.992 & 19857.270 & 21265.656 & 37503.782 & \textbf{13772.419} & \textbf{19.08} \\
Z200 & \underline{18855.467} & 24490.470 & 19590.327 & 21580.535 & 22028.481 & 34481.412 & \textbf{15841.694} & \textbf{15.98} \\
Z250 & \underline{20104.820} & 24402.604 & 20700.654 & 22090.788 & 22650.018 & 32376.043 & \textbf{17707.925} & \textbf{11.92} \\
Z300 & \underline{19388.488} & 22716.411 & 19957.603 & 21041.784 & 21442.618 & 28922.522 & \textbf{17547.251} & \textbf{9.50} \\
Z400 & \underline{14892.992} & 16872.329 & 15389.543 & 15762.731 & 16319.090 & 20502.533 & \textbf{13792.311} & \textbf{7.39} \\
Z500 & \underline{11389.439} & 12592.577 & 11796.738 & 11940.985 & 12350.254 & 14763.485 & \textbf{10613.034} & \textbf{6.82} \\
\end{longtable}
\endgroup
\clearpage

\begingroup
\scriptsize
\setlength{\tabcolsep}{3pt}
\renewcommand{\arraystretch}{0.95}
\begin{longtable}{lrrrrrrrr}
\caption{MAE for all 69 output fields at a 24-hour lead time. Bold and underline indicate the lowest and second-lowest values, respectively. Gain is the relative MAE reduction of Ours over the best baseline for each field; negative values indicate that Ours has a larger error.}\label{tab:all69_mae_24h_seven_methods} \\
\toprule
Field & $\delta$-Adapter & Proceed & OneNet & FSNet & ADAPT-Z & ELF & Ours & Gain (\%) \\
\midrule
\endfirsthead
\multicolumn{9}{c}{\tablename\ \thetable{} -- continued} \\
\toprule
Field & $\delta$-Adapter & Proceed & OneNet & FSNet & ADAPT-Z & ELF & Ours & Gain (\%) \\
\midrule
\endhead
\midrule
\multicolumn{9}{r}{Continued on next page} \\
\endfoot
\bottomrule
\endlastfoot
Z50 & \underline{118.743} & 141.393 & 124.028 & 128.305 & 131.246 & 159.717 & \textbf{101.165} & \textbf{14.80} \\
Z100 & \underline{95.434} & 118.877 & 101.260 & 106.527 & 109.220 & 135.536 & \textbf{82.462} & \textbf{13.59} \\
Z150 & \underline{95.775} & 117.858 & 99.352 & 105.778 & 108.911 & 144.822 & \textbf{83.760} & \textbf{12.54} \\
Z200 & \underline{98.123} & 117.053 & 100.677 & 108.599 & 108.811 & 143.980 & \textbf{87.047} & \textbf{11.29} \\
Z250 & \underline{98.184} & 114.108 & 100.103 & 106.729 & 107.439 & 137.687 & \textbf{89.151} & \textbf{9.20} \\
Z300 & \underline{94.338} & 107.955 & 96.205 & 102.119 & 102.269 & 128.460 & \textbf{86.859} & \textbf{7.93} \\
Z400 & \underline{81.093} & 90.956 & 83.029 & 86.104 & 87.522 & 106.177 & \textbf{75.871} & \textbf{6.44} \\
Z500 & \underline{70.940} & 77.824 & 72.701 & 74.606 & 75.807 & 88.890 & \textbf{66.748} & \textbf{5.91} \\
\end{longtable}
\endgroup
\clearpage

\begingroup
\scriptsize
\setlength{\tabcolsep}{3pt}
\renewcommand{\arraystretch}{0.95}
\begin{longtable}{lrrrrrrrr}
\caption{MSE for all 69 output fields at a 48-hour lead time. Bold and underline indicate the lowest and second-lowest values, respectively. Gain is the relative MSE reduction of Ours over the best baseline for each field; negative values indicate that Ours has a larger error.}\label{tab:all69_mse_48h_seven_methods} \\
\toprule
Field & $\delta$-Adapter & Proceed & OneNet & FSNet & ADAPT-Z & ELF & Ours & Gain (\%) \\
\midrule
\endfirsthead
\multicolumn{9}{c}{\tablename\ \thetable{} -- continued} \\
\toprule
Field & $\delta$-Adapter & Proceed & OneNet & FSNet & ADAPT-Z & ELF & Ours & Gain (\%) \\
\midrule
\endhead
\midrule
\multicolumn{9}{r}{Continued on next page} \\
\endfoot
\bottomrule
\endlastfoot
Z50 & \underline{68812.226} & 89114.694 & 75692.650 & 77697.253 & 83445.627 & 112711.138 & \textbf{58733.177} & \textbf{14.65} \\
Z100 & \underline{44757.219} & 59551.570 & 49718.579 & 51682.857 & 55013.873 & 86097.088 & \textbf{37863.339} & \textbf{15.40} \\
Z150 & \underline{50015.601} & 62731.708 & 52945.670 & 55696.941 & 58191.989 & 93854.958 & \textbf{43563.154} & \textbf{12.90} \\
Z200 & \underline{60880.093} & 70775.338 & 62955.439 & 65752.393 & 66834.789 & 89626.985 & \textbf{54928.430} & \textbf{9.78} \\
Z250 & \underline{70460.119} & 77784.912 & 72221.423 & 73771.876 & 75202.398 & 92636.280 & \textbf{65853.003} & \textbf{6.54} \\
Z300 & \underline{70995.989} & 76544.662 & 72711.675 & 73530.367 & 74920.230 & 87697.973 & \textbf{67557.670} & \textbf{4.84} \\
Z400 & \underline{55518.409} & 58743.690 & 56986.463 & 56736.475 & 58309.483 & 64965.811 & \textbf{53531.662} & \textbf{3.58} \\
Z500 & \underline{41063.557} & 42963.359 & 42205.014 & 41778.181 & 43002.050 & 46638.255 & \textbf{39626.369} & \textbf{3.50} \\
\end{longtable}
\endgroup
\clearpage

\begingroup
\scriptsize
\setlength{\tabcolsep}{3pt}
\renewcommand{\arraystretch}{0.95}
\begin{longtable}{lrrrrrrrr}
\caption{MAE for all 69 output fields at a 48-hour lead time. Bold and underline indicate the lowest and second-lowest values, respectively. Gain is the relative MAE reduction of Ours over the best baseline for each field; negative values indicate that Ours has a larger error.}\label{tab:all69_mae_48h_seven_methods} \\
\toprule
Field & $\delta$-Adapter & Proceed & OneNet & FSNet & ADAPT-Z & ELF & Ours & Gain (\%) \\
\midrule
\endfirsthead
\multicolumn{9}{c}{\tablename\ \thetable{} -- continued} \\
\toprule
Field & $\delta$-Adapter & Proceed & OneNet & FSNet & ADAPT-Z & ELF & Ours & Gain (\%) \\
\midrule
\endhead
\midrule
\multicolumn{9}{r}{Continued on next page} \\
\endfoot
\bottomrule
\endlastfoot
Z50 & \underline{178.619} & 212.370 & 186.219 & 193.602 & 200.091 & 247.914 & \textbf{155.990} & \textbf{12.67} \\
Z100 & \underline{150.908} & 181.499 & 159.593 & 166.625 & 170.897 & 206.615 & \textbf{133.830} & \textbf{11.32} \\
Z150 & \underline{158.488} & 185.849 & 164.471 & 172.926 & 176.003 & 222.310 & \textbf{143.903} & \textbf{9.20} \\
Z200 & \underline{168.741} & 191.621 & 173.226 & 182.484 & 182.655 & 225.647 & \textbf{156.106} & \textbf{7.49} \\
Z250 & \underline{174.021} & 192.761 & 177.413 & 184.698 & 185.745 & 221.968 & \textbf{164.372} & \textbf{5.54} \\
Z300 & \underline{169.648} & 185.352 & 172.909 & 178.663 & 179.595 & 210.225 & \textbf{162.010} & \textbf{4.50} \\
Z400 & \underline{146.370} & 157.395 & 149.531 & 151.712 & 154.116 & 175.585 & \textbf{141.189} & \textbf{3.54} \\
Z500 & \underline{126.361} & 133.788 & 129.064 & 129.773 & 132.118 & 146.970 & \textbf{122.008} & \textbf{3.45} \\
\end{longtable}
\endgroup
\clearpage

\begingroup
\scriptsize
\setlength{\tabcolsep}{3pt}
\renewcommand{\arraystretch}{0.95}
\begin{longtable}{lrrrrrrrr}
\caption{MSE for all 69 output fields at a 72-hour lead time. Bold and underline indicate the lowest and second-lowest values, respectively. Gain is the relative MSE reduction of Ours over the best baseline for each field; negative values indicate that Ours has a larger error.}\label{tab:all69_mse_72h_seven_methods} \\
\toprule
Field & $\delta$-Adapter & Proceed & OneNet & FSNet & ADAPT-Z & ELF & Ours & Gain (\%) \\
\midrule
\endfirsthead
\multicolumn{9}{c}{\tablename\ \thetable{} -- continued} \\
\toprule
Field & $\delta$-Adapter & Proceed & OneNet & FSNet & ADAPT-Z & ELF & Ours & Gain (\%) \\
\midrule
\endhead
\midrule
\multicolumn{9}{r}{Continued on next page} \\
\endfoot
\bottomrule
\endlastfoot
Z50 & \underline{127630.585} & 157641.646 & 139631.404 & 140705.027 & 154219.067 & 194965.406 & \textbf{114226.842} & \textbf{10.50} \\
Z100 & \underline{88607.526} & 108506.637 & 97391.056 & 98428.520 & 104697.355 & 151564.655 & \textbf{79194.299} & \textbf{10.62} \\
Z150 & \underline{105769.800} & 122059.215 & 110973.460 & 113858.677 & 117057.361 & 168633.847 & \textbf{96760.022} & \textbf{8.52} \\
Z200 & \underline{136764.726} & 149241.663 & 140463.066 & 143324.937 & 144575.951 & 174454.667 & \textbf{128267.043} & \textbf{6.21} \\
Z250 & \underline{165035.910} & 174096.839 & 168342.360 & 169374.636 & 171109.688 & 194112.624 & \textbf{158221.614} & \textbf{4.13} \\
Z300 & \underline{168433.735} & 175121.396 & 171627.696 & 171483.450 & 173396.785 & 189888.639 & \textbf{163163.223} & \textbf{3.13} \\
Z400 & \underline{130863.149} & 134552.518 & 133469.039 & 132087.359 & 134346.056 & 142558.114 & \textbf{127744.759} & \textbf{2.38} \\
Z500 & \underline{94418.273} & 96396.175 & 96326.891 & 95054.908 & 96828.131 & 101070.648 & \textbf{92146.142} & \textbf{2.41} \\
\end{longtable}
\endgroup
\clearpage

\begingroup
\scriptsize
\setlength{\tabcolsep}{3pt}
\renewcommand{\arraystretch}{0.95}
\begin{longtable}{lrrrrrrrr}
\caption{MAE for all 69 output fields at a 72-hour lead time. Bold and underline indicate the lowest and second-lowest values, respectively. Gain is the relative MAE reduction of Ours over the best baseline for each field; negative values indicate that Ours has a larger error.}\label{tab:all69_mae_72h_seven_methods} \\
\toprule
Field & $\delta$-Adapter & Proceed & OneNet & FSNet & ADAPT-Z & ELF & Ours & Gain (\%) \\
\midrule
\endfirsthead
\multicolumn{9}{c}{\tablename\ \thetable{} -- continued} \\
\toprule
Field & $\delta$-Adapter & Proceed & OneNet & FSNet & ADAPT-Z & ELF & Ours & Gain (\%) \\
\midrule
\endhead
\midrule
\multicolumn{9}{r}{Continued on next page} \\
\endfoot
\bottomrule
\endlastfoot
Z50 & \underline{230.963} & 269.933 & 240.091 & 248.546 & 259.298 & 316.863 & \textbf{205.915} & \textbf{10.85} \\
Z100 & \underline{205.115} & 237.709 & 215.530 & 223.400 & 229.036 & 267.376 & \textbf{186.958} & \textbf{8.85} \\
Z150 & \underline{223.690} & 251.878 & 231.169 & 240.753 & 243.103 & 291.668 & \textbf{209.195} & \textbf{6.48} \\
Z200 & \underline{244.447} & 267.750 & 250.120 & 259.968 & 259.493 & 303.395 & \textbf{232.414} & \textbf{4.92} \\
Z250 & \underline{256.870} & 275.788 & 261.302 & 268.739 & 269.383 & 306.413 & \textbf{247.945} & \textbf{3.47} \\
Z300 & \underline{252.148} & 267.820 & 256.338 & 261.763 & 262.672 & 293.768 & \textbf{245.182} & \textbf{2.76} \\
Z400 & \underline{217.321} & 228.068 & 221.183 & 222.663 & 225.325 & 247.076 & \textbf{212.673} & \textbf{2.14} \\
Z500 & \underline{185.546} & 192.539 & 188.721 & 188.621 & 191.408 & 206.438 & \textbf{181.575} & \textbf{2.14} \\
\end{longtable}
\endgroup
\clearpage
\subsection{Gating weight results}
\label{Gating weight}
\begin{figure}[htbp]
    \centering
    \includegraphics[width=1.0\linewidth]{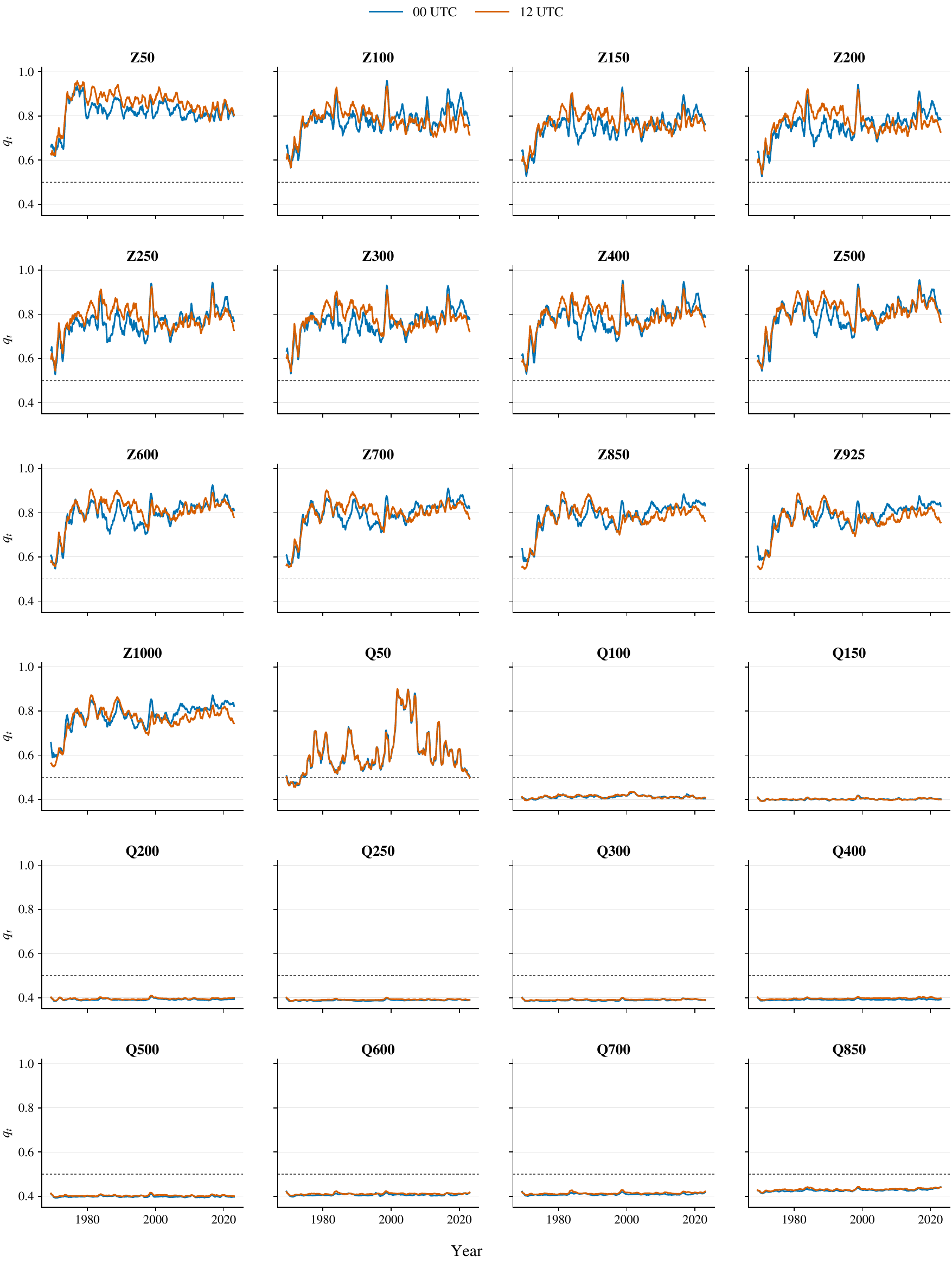}
    \caption{
Temporal evolution of the routing weights for geopotential
($\mathrm{Z50}$--$\mathrm{Z1000}$) and upper-level specific humidity
($\mathrm{Q50}$--$\mathrm{Q850}$). Blue and orange curves show the
365-day rolling means for the 00 and 12 UTC initialization streams,
respectively.
}
    \label{fig:g1}
\end{figure}
\begin{figure}[htbp]
    \centering
    \includegraphics[width=1.0\linewidth]{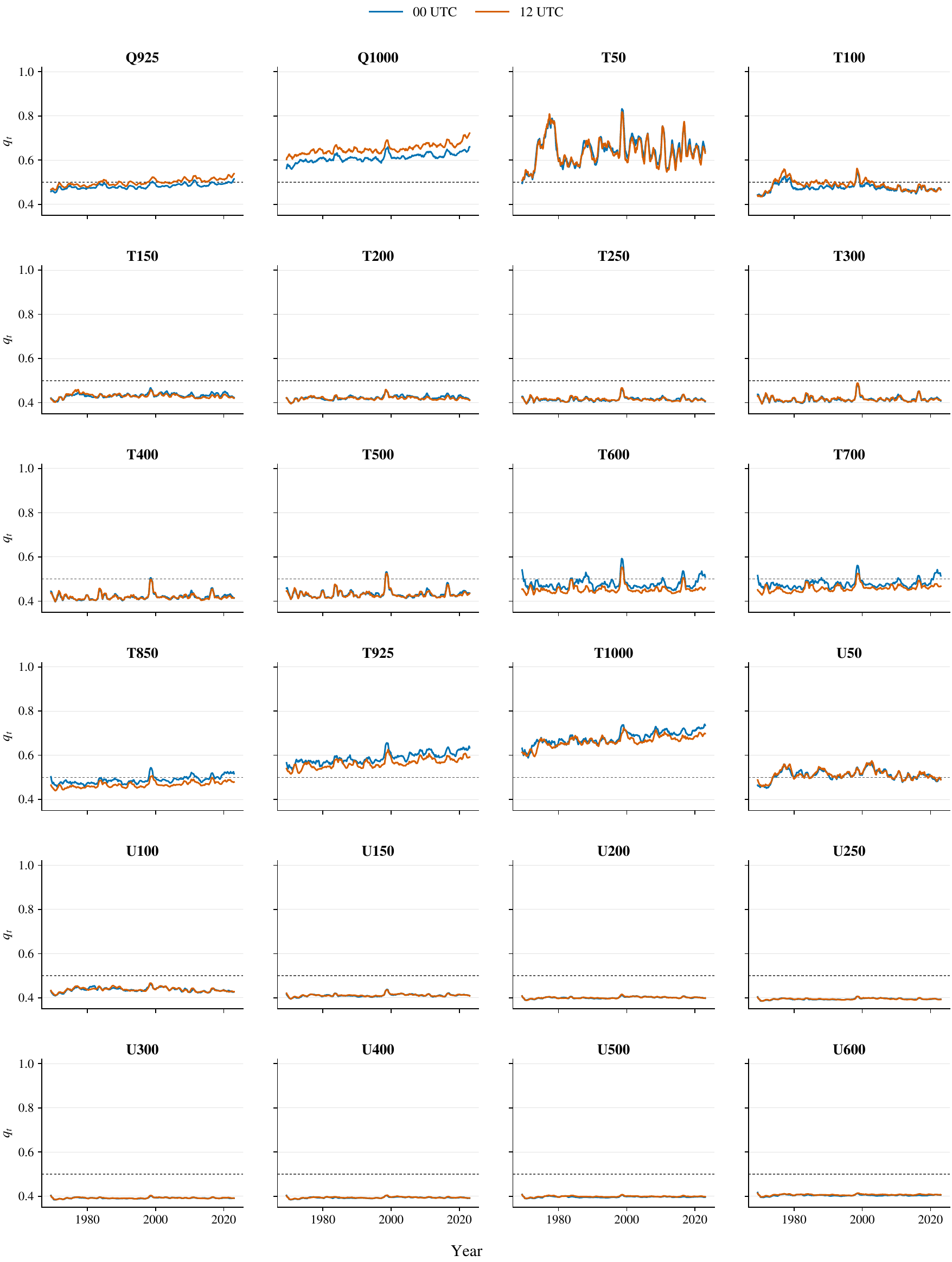}
    \caption{
Temporal evolution of the routing weights for lower-level specific
humidity ($\mathrm{Q925}$ and $\mathrm{Q1000}$), temperature
($\mathrm{T50}$--$\mathrm{T1000}$), and upper-level zonal wind
($\mathrm{U50}$--$\mathrm{U600}$). Blue and orange curves show the
365-day rolling means for the 00 and 12 UTC initialization streams,
respectively. 
}
    \label{g2}
\end{figure}
\begin{figure}[htbp]
    \centering
    \includegraphics[width=1.0\linewidth]{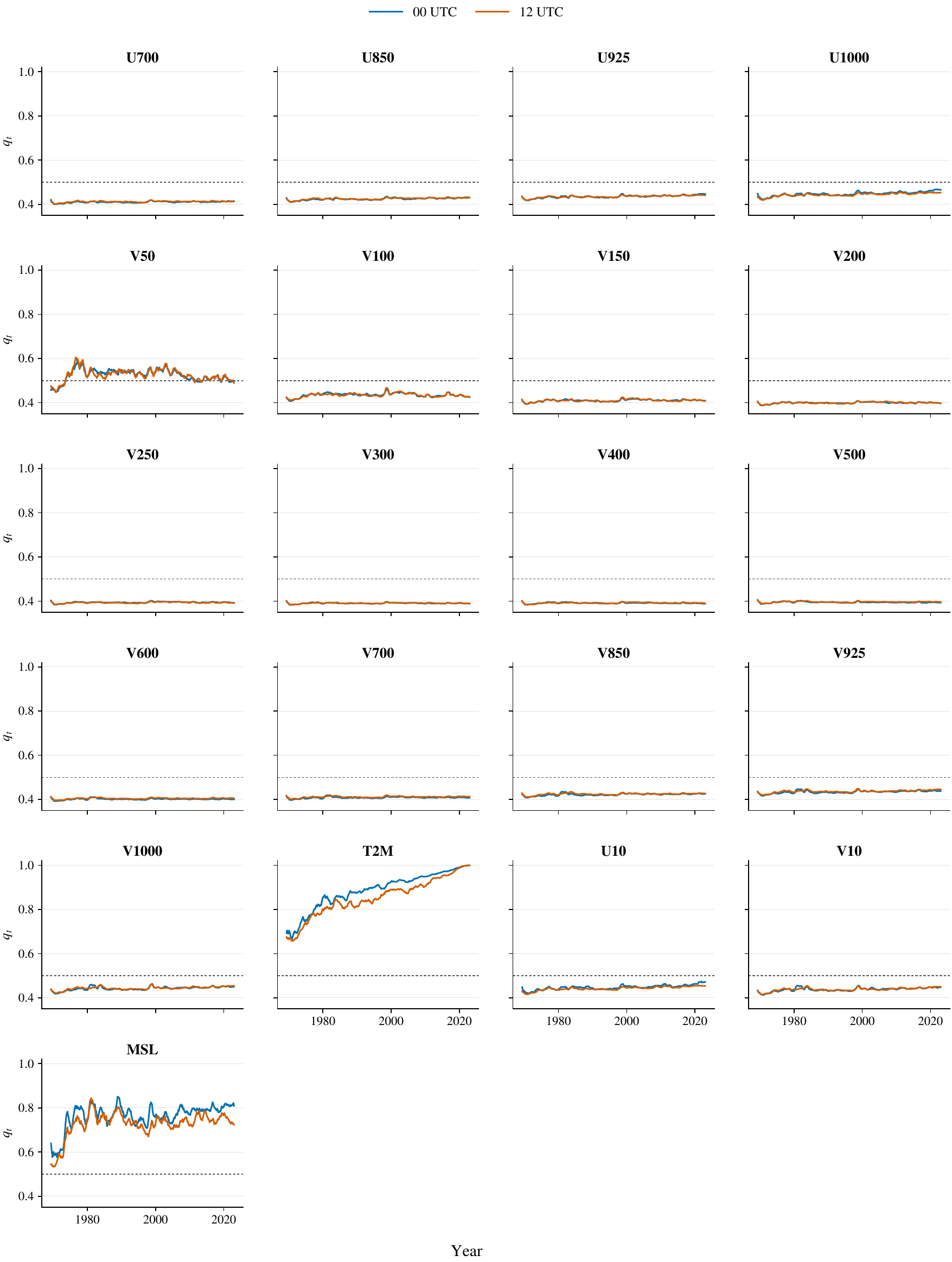}
    \caption{
Temporal evolution of the routing weights for lower-level zonal wind
($\mathrm{U700}$--$\mathrm{U1000}$), meridional wind
($\mathrm{V50}$--$\mathrm{V1000}$), and surface fields
(T2M, U10, V10, and MSL). Blue and orange curves show the 365-day
rolling means for the 00 and 12 UTC initialization streams,
respectively.
}
    \label{g3}
\end{figure}
\clearpage
\subsection{Case-study visualizations}
\label{Case-study visualizations}
\begin{figure}[htbp]
    \centering
    \includegraphics[width=1.0\linewidth]{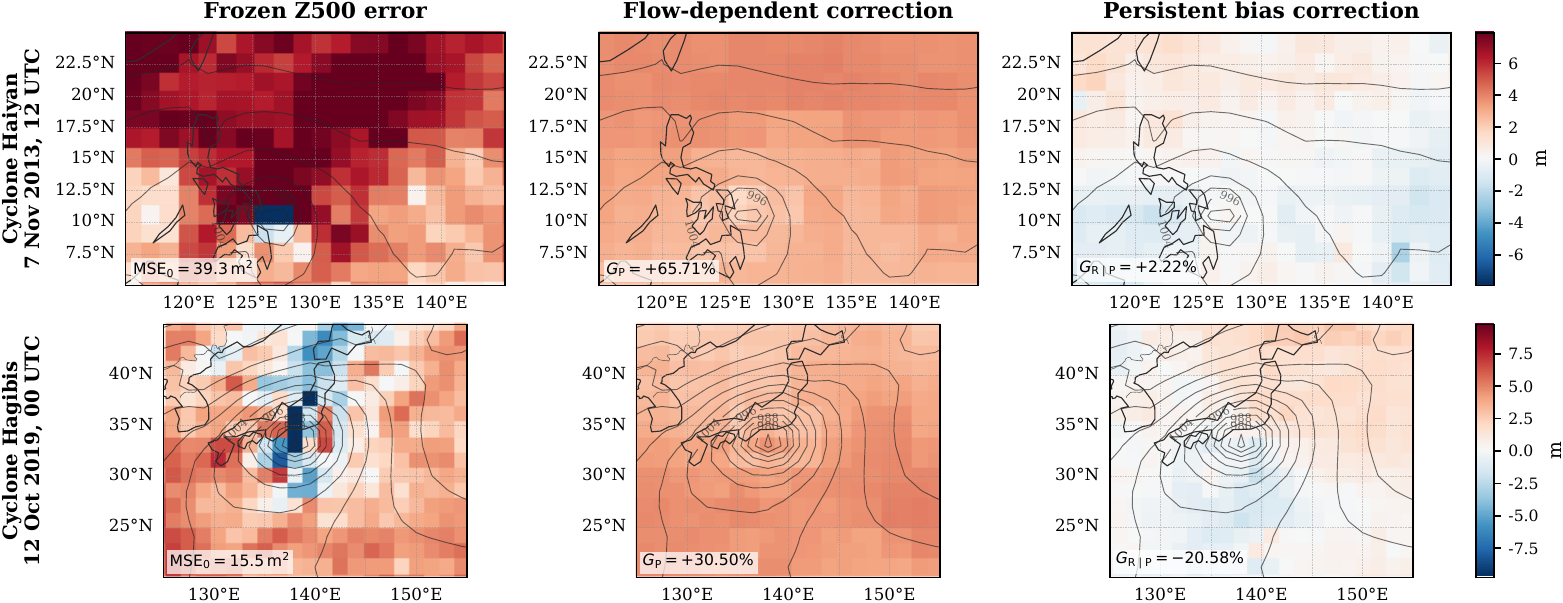}
    \caption{
\textbf{Decomposition of Z500 forecast-error correction for two tropical-cyclone cases.}
Rows show forecasts initialized during Typhoon Haiyan (7 November 2013, 12 UTC) and Typhoon Hagibis (12 October 2019, 00 UTC).
The frozen-model MSE and the relative contributions of the flow-dependent ($G_{\mathrm{P}}$) and persistent ($G_{\mathrm{R}\mid\mathrm{P}}$) corrections are reported in each panel.
}
    \label{fig:typhoon}
\end{figure}
\subsection{Variable-wise Climate-Shift and Adaptation Analysis}
\label{A:variable_wise_climate_shift}
To examine whether the temporal-drift behavior observed for T2M extends
to other atmospheric variables, we repeat the analysis for Z500, T850,
U850, V850, U10, V10, and MSL.

\begin{figure}[htbp]
    \centering
    \includegraphics[width=1\linewidth]{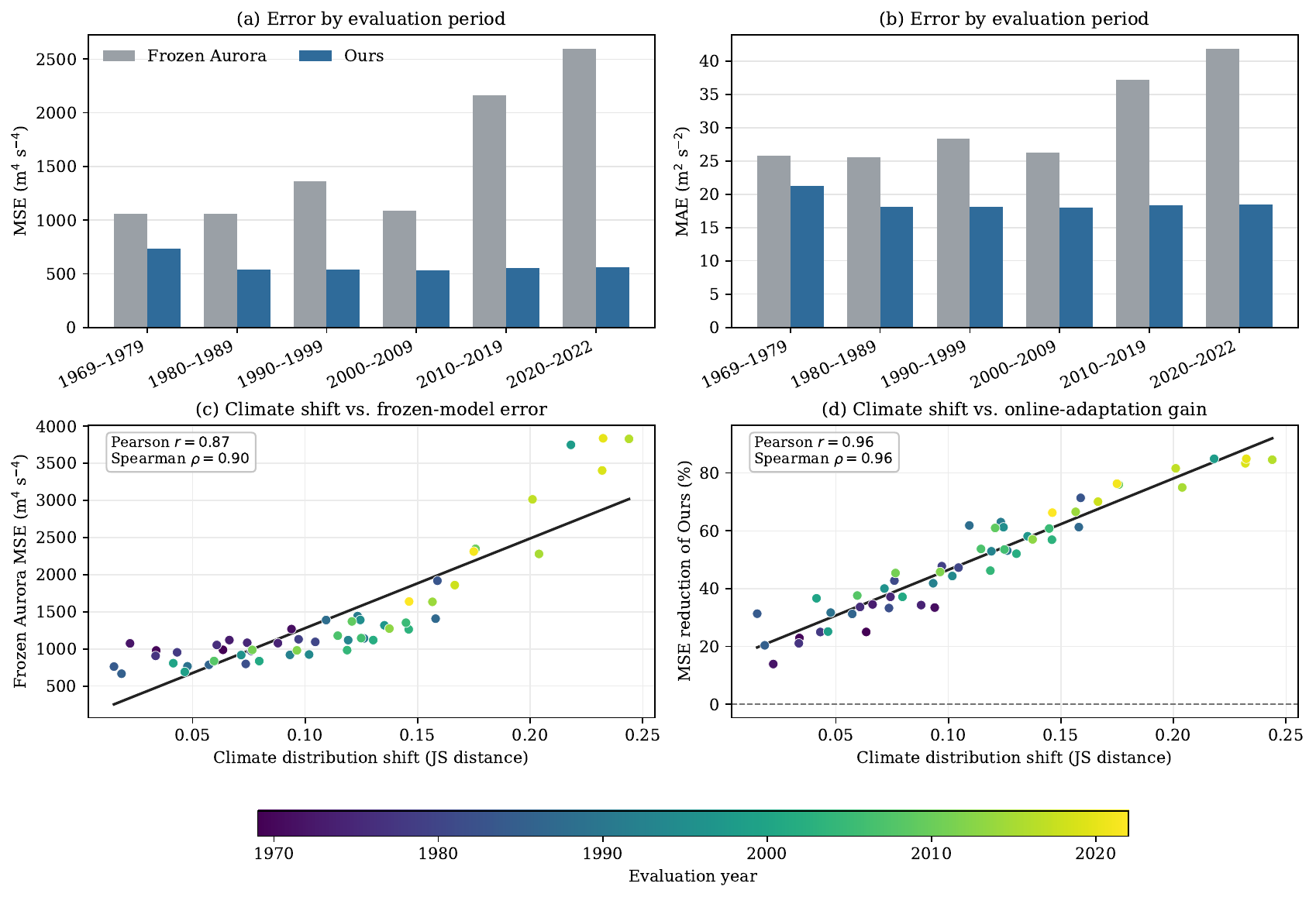}
    \caption{Climate-shift analysis for Z500.}
    \label{fig:z500_changing_climate}
\end{figure}

\begin{figure}[htbp]
    \centering
    \includegraphics[width=1\linewidth]{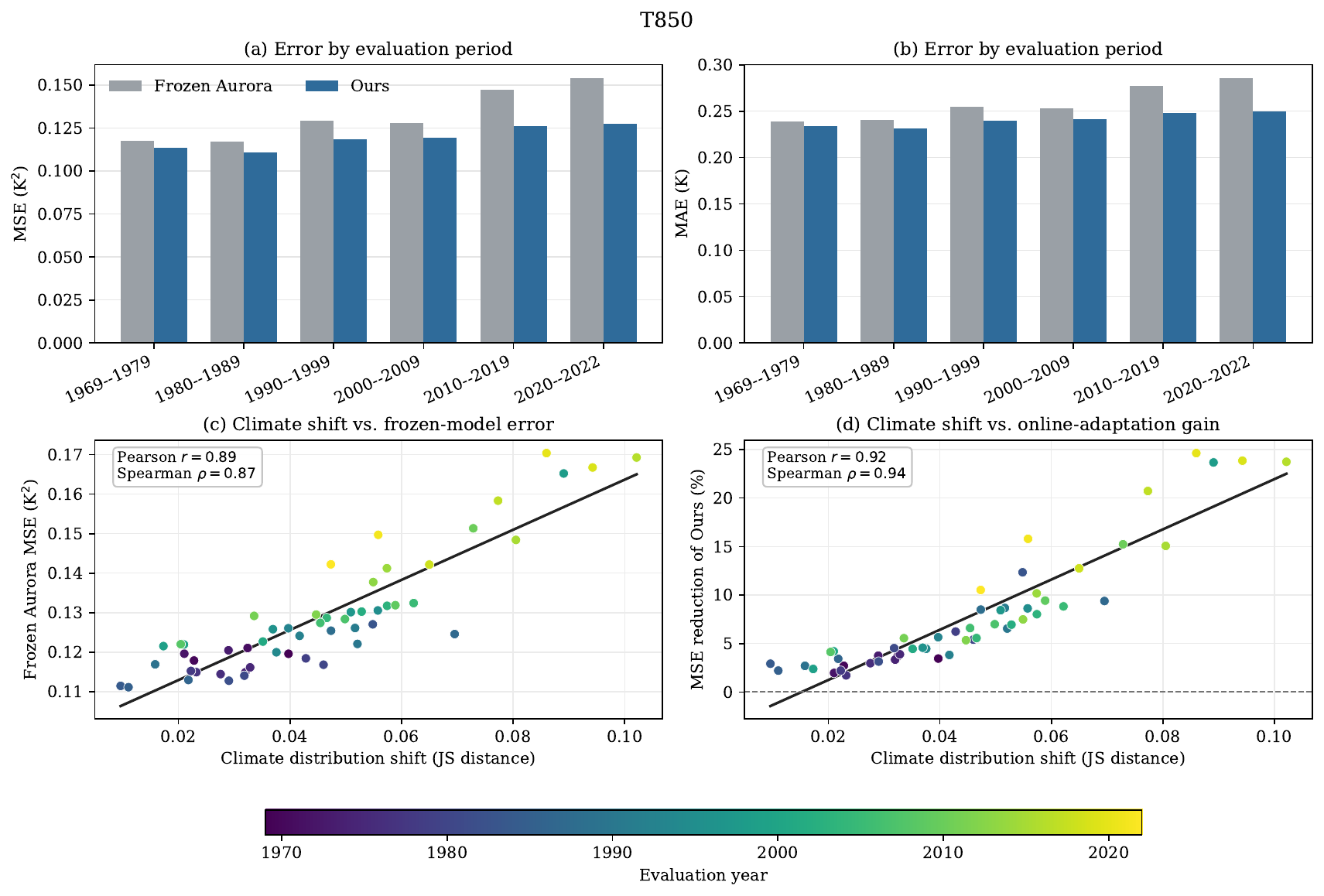}
    \caption{Climate-shift analysis for T850.}
    \label{fig:t850_changing_climate}
\end{figure}

Figures~\ref{fig:z500_changing_climate} and
\ref{fig:t850_changing_climate} show the clearest temporal degradation.
For Z500, climate shift is strongly associated with both the MSE of
Frozen Aurora ($r=0.87$) and the MSE reduction achieved by our method
($r=0.96$). T850 exhibits similarly strong relationships, with
$r=0.89$ for frozen-model error and $r=0.92$ for adaptation gain. The
period-averaged results further show that the errors of Frozen Aurora
increase markedly in the later evaluation periods, whereas online
adaptation substantially limits this degradation.

\begin{figure}[htbp]
    \centering
    \includegraphics[width=1\linewidth]{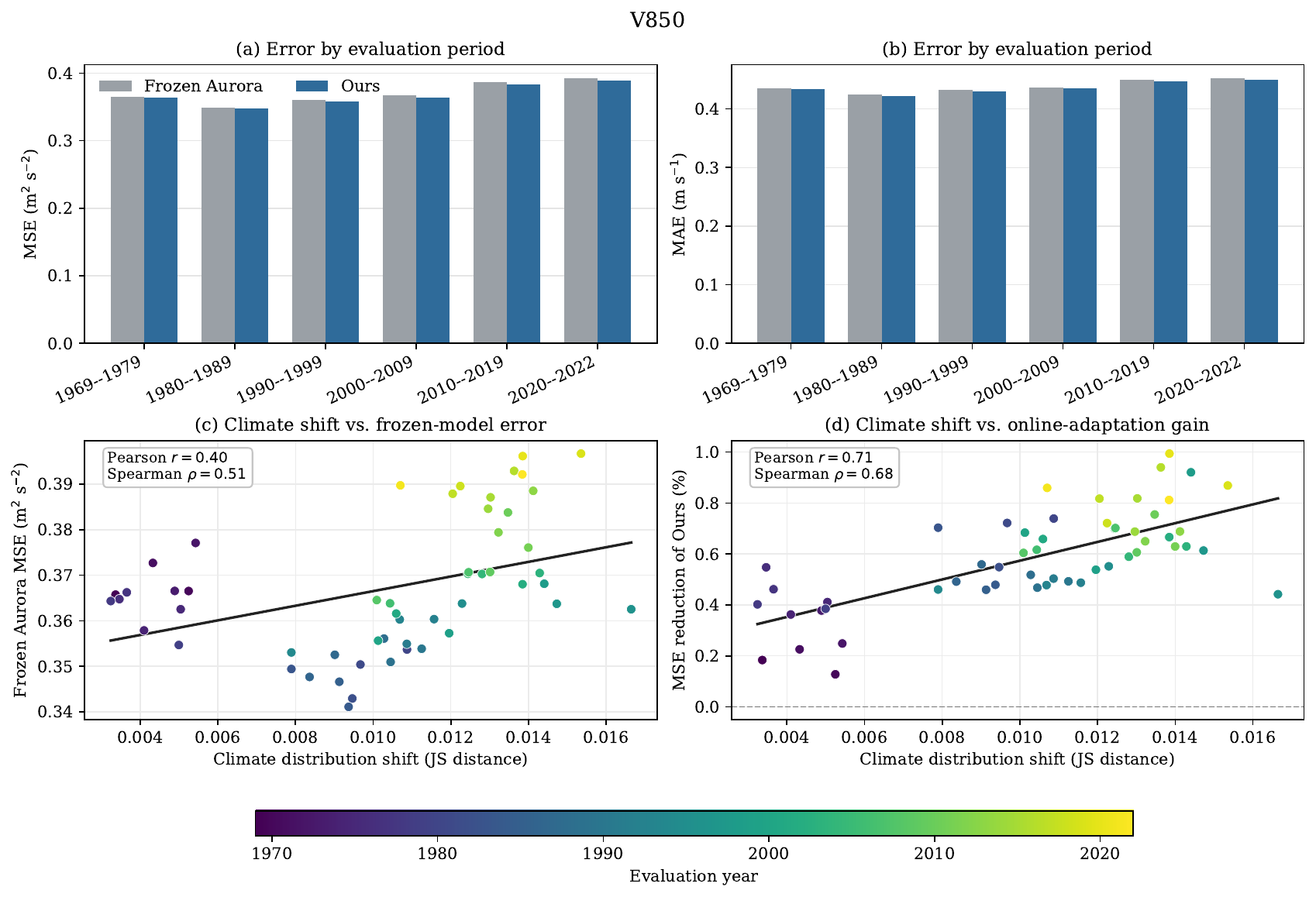}
    \caption{Climate-shift analysis for V850.}
    \label{fig:v850_changing_climate}
\end{figure}

\begin{figure}[htbp]
    \centering
    \includegraphics[width=1\linewidth]{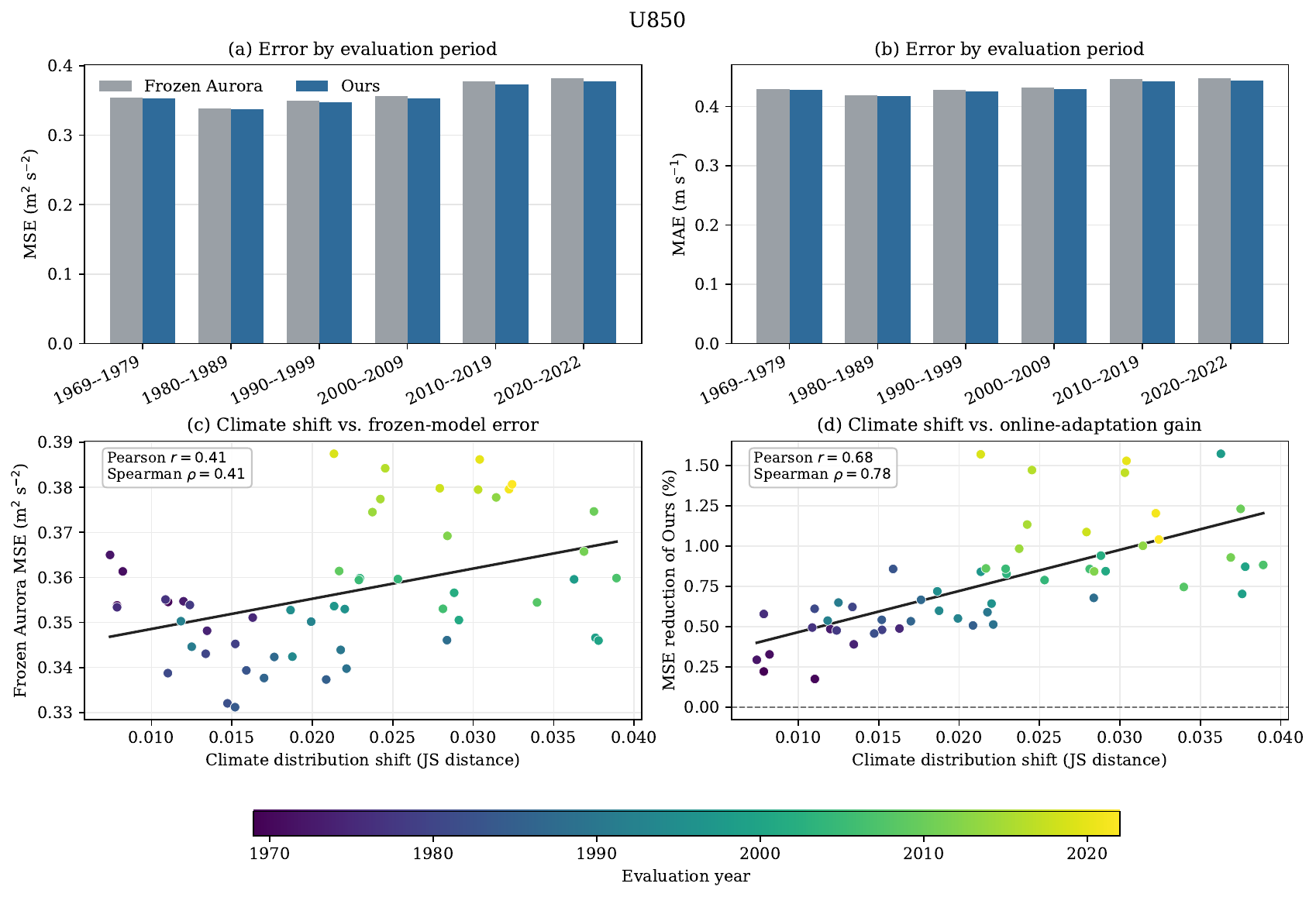}
    \caption{Climate-shift analysis for U850.}
    \label{fig:u850_changing_climate}
\end{figure}

\begin{figure}[htbp]
    \centering
    \includegraphics[width=1\linewidth]{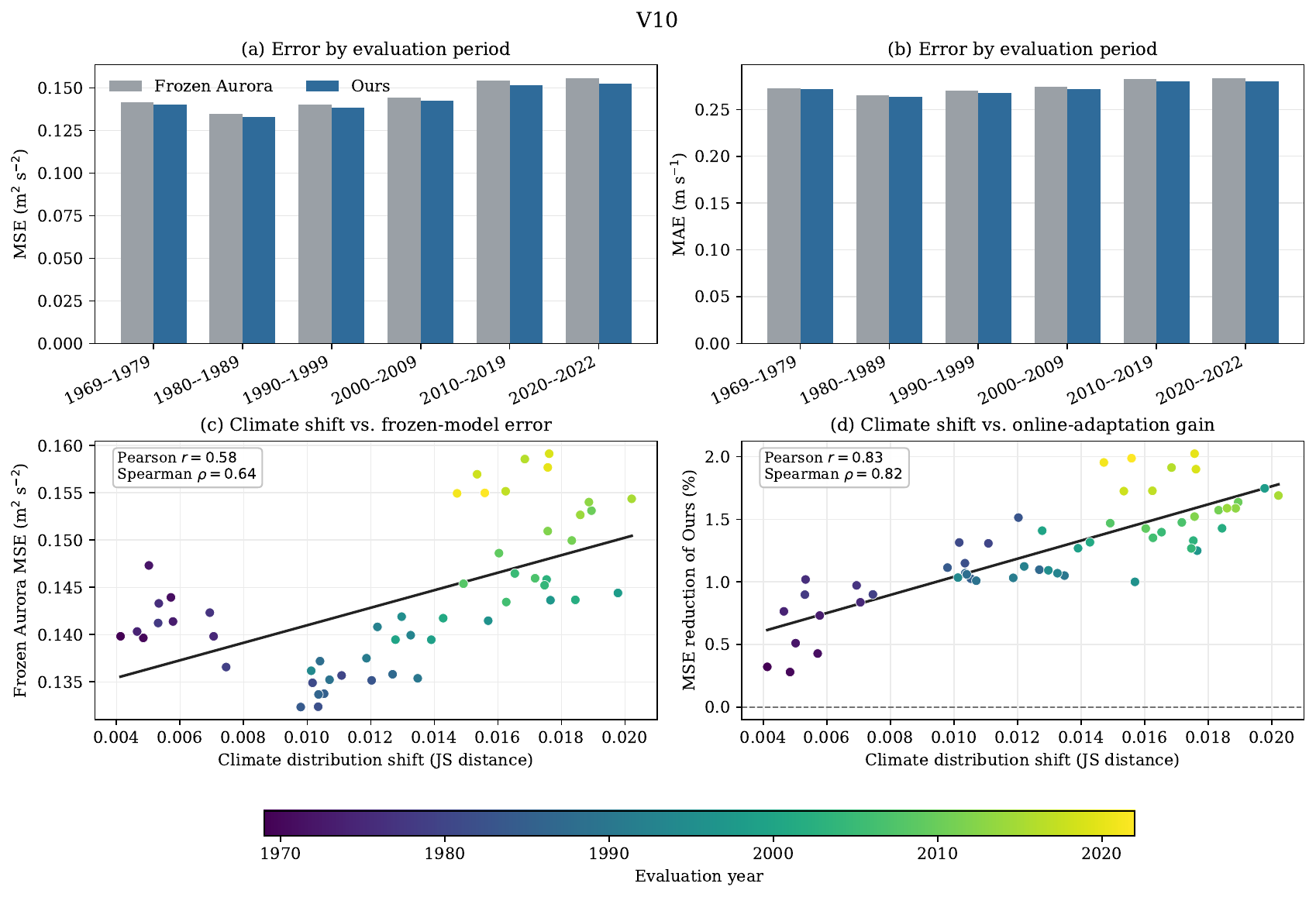}
    \caption{Climate-shift analysis for V10.}
    \label{fig:v10_changing_climate}
\end{figure}

\begin{figure}[htbp]
    \centering
    \includegraphics[width=1\linewidth]{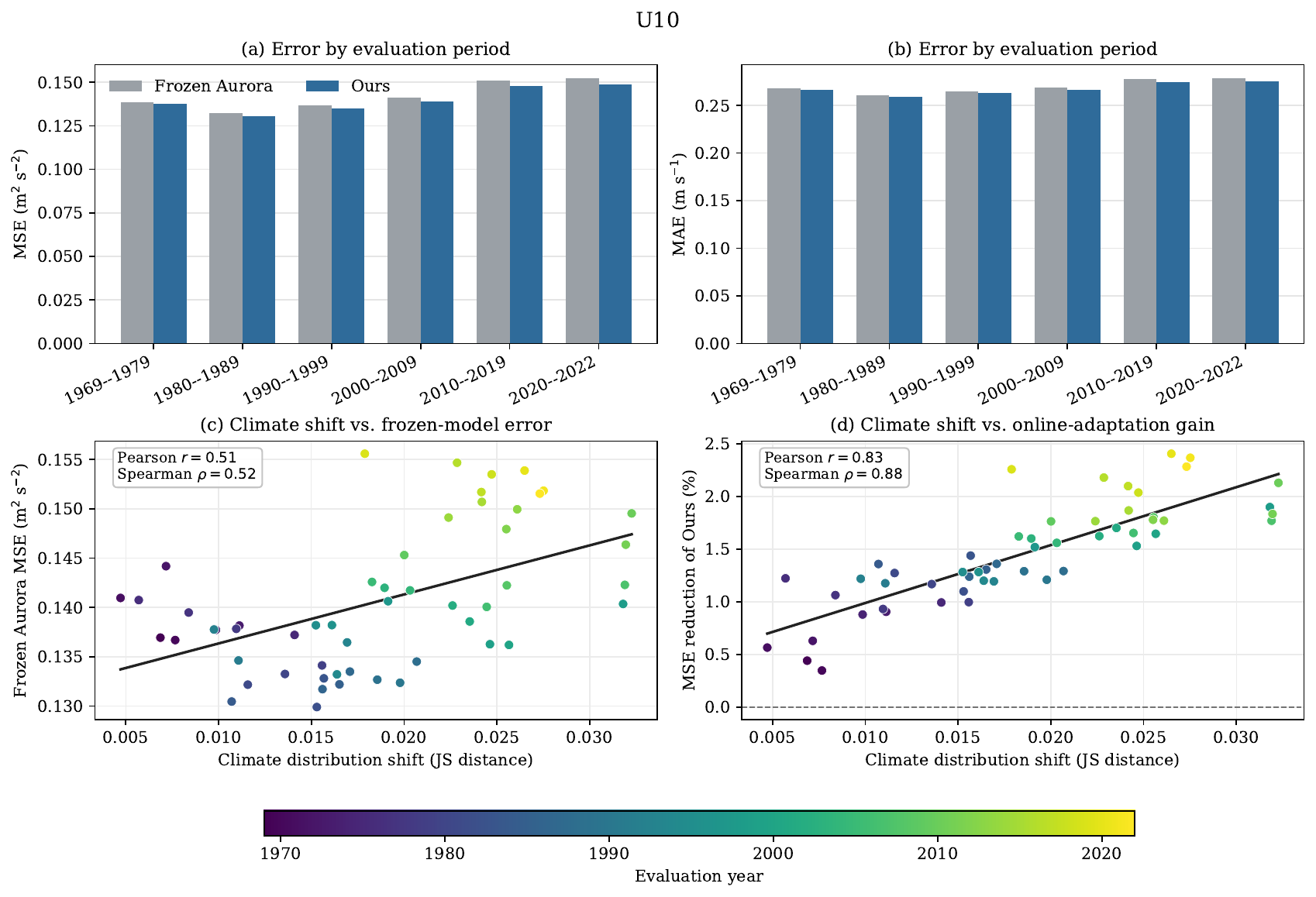}
    \caption{Climate-shift analysis for U10.}
    \label{fig:u10_changing_climate}
\end{figure}

The wind variables exhibit weaker relationships between climate shift
and frozen-model error, but more consistent relationships between shift
and adaptation gain. As shown in
Figures~\ref{fig:u850_changing_climate},
\ref{fig:v850_changing_climate},
\ref{fig:u10_changing_climate}, and
\ref{fig:v10_changing_climate}, the correlations with frozen-model MSE
range from $0.40$ to $0.58$, while the correlations with adaptation gain
range from $0.68$ to $0.83$. Although the absolute improvements for the
wind fields are smaller than those for Z500 and T850, the increasing
adaptation gains indicate that online updating becomes more useful as
their deployment distributions move farther from the training climate.

\begin{figure}[htbp]
    \centering
    \includegraphics[width=1\linewidth]{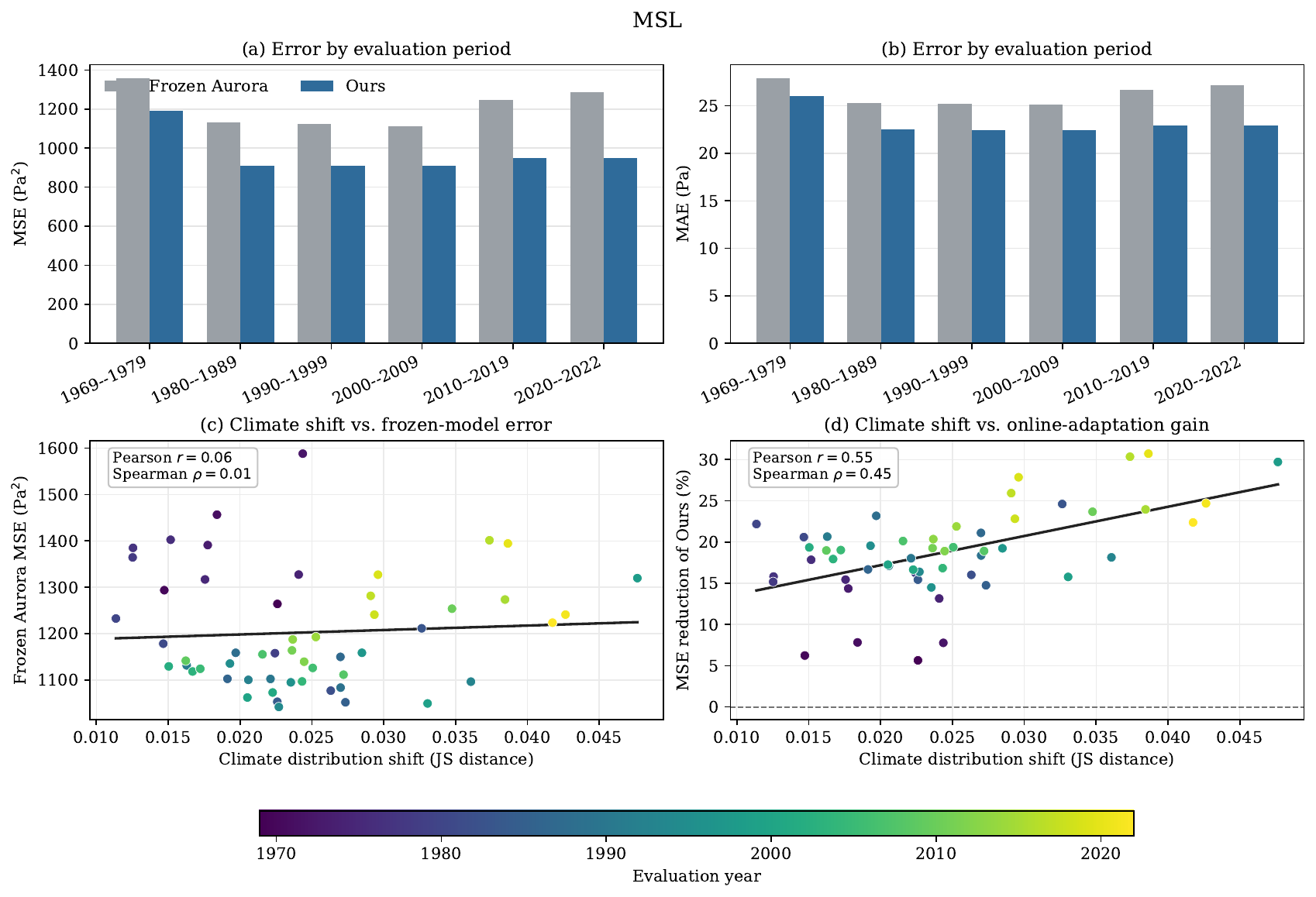}
    \caption{Climate-shift analysis for MSL.}
    \label{fig:msl_changing_climate}
\end{figure}

MSL presents a different pattern
(Figure~\ref{fig:msl_changing_climate}). Its climate distribution shift
has almost no association with frozen-model error ($r=0.06$), although
it remains moderately associated with adaptation gain ($r=0.55$).
Unlike temperature and geopotential, MSL errors depend strongly on the
spatial organization and displacement of synoptic pressure systems.
Consequently, a global marginal-distribution distance may not fully
capture the circulation-related changes that govern MSL forecast
errors. This result highlights that the relationship between climate
shift and forecast degradation is variable dependent and that no single
marginal drift measure is expected to describe every atmospheric field
equally well.

Overall, the additional variables support the broader conclusion that
forecast degradation and the benefit of online adaptation generally
increase under stronger distribution shift. However, the magnitude of
this relationship depends on the physical characteristics of each
variable and on whether the selected drift measure captures its
relevant spatial and dynamical structure.

\clearpage
\subsection{Ablation of the two correction components}
\label{A:Ablation of the two correction components}
\begin{figure}[h]
    \centering
    \includegraphics[width=1.0\linewidth]{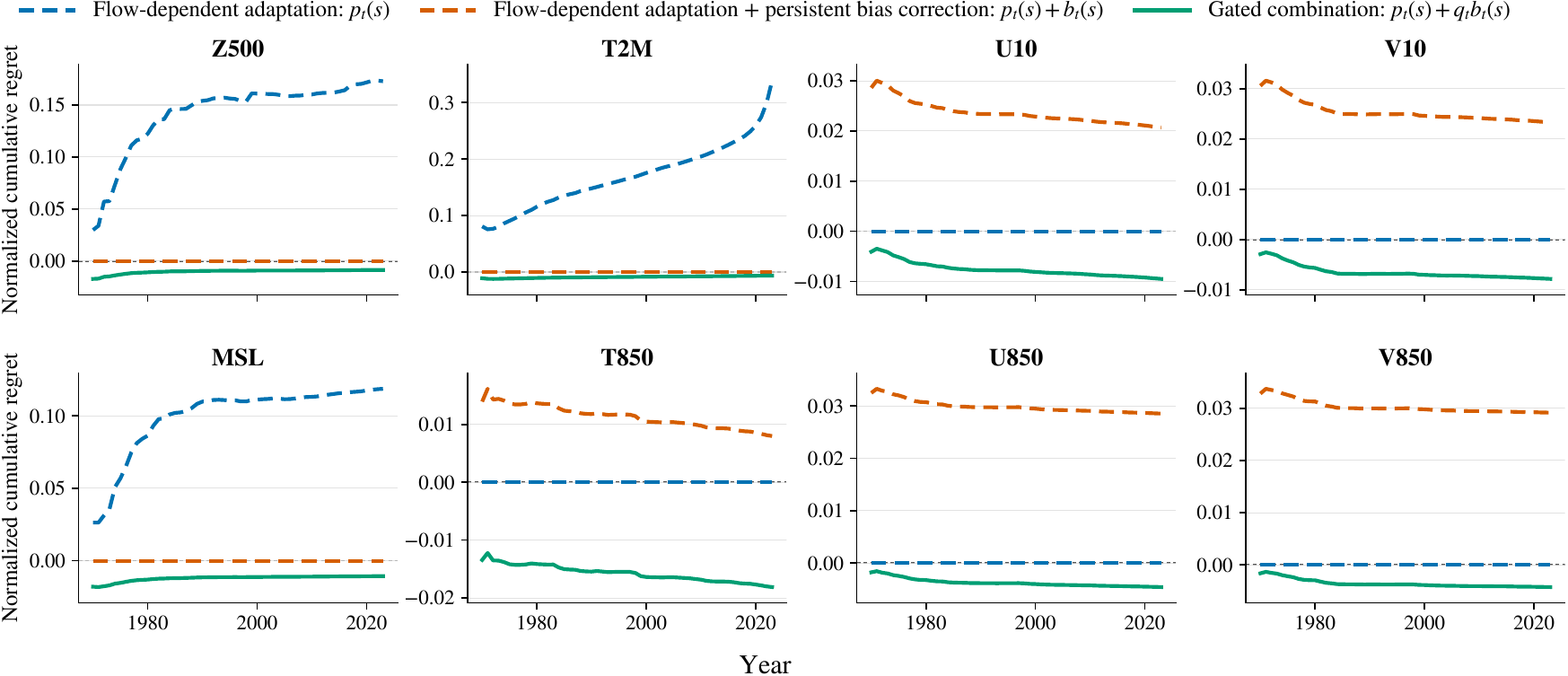}
    \caption{Normalized cumulative regret of the adaptation endpoints and routed
forecast relative to the best fixed endpoint in hindsight.
Negative regret indicates that the routed mixture outperforms the best
fixed endpoint over the corresponding period.}
    \label{fig:endpoint_regret}
\end{figure}
\end{document}